\documentclass[10pt,twocolumn,letterpaper]{article}
\usepackage{xcolor}
\definecolor{cvprblue}{rgb}{0.21,0.49,0.74}

\newcommand{\mytilde}{\raise.17ex\hbox{$\scriptstyle\mathtt{\sim}$}}
\usepackage[pagenumbers]{cvpr} % To force page numbers, e.g. for an arXiv version
\usepackage{soul}
\usepackage{algorithm}
\usepackage{algpseudocode}

\usepackage[pagebackref,breaklinks,colorlinks,allcolors=cvprblue]{hyperref}

\def\paperID{2835} % *** Enter the Paper ID here
\def\confName{CVPR}
\def\confYear{2025}

\newcommand{\methodname}{UVGS\xspace}

\title{\methodname: Reimagining Unstructured 3D Gaussian Splatting using UV Mapping}

\vspace{-0.6cm}

\author{
Aashish Rai$^{1,2}$
\and
Dilin Wang$^{2}$
\and
Mihir Jain$^{2}$
\and
Nikolaos Sarafianos$^{2}$
\and
Kefan Chen$^{1,2}$
\and
Srinath Sridhar$^1$
\and
Aayush Prakash$^2$
\vspace{0.05in}
\and
\centerline{$^1$Brown University \hspace{0.2in} $^2$Meta Reality Labs}
\vspace{0.01in}
\and
{\tt\small \url{https://ivl.cs.brown.edu/uvgs}}
}

\begin{document}
\twocolumn[{%
\renewcommand\twocolumn[1][]{#1}%
\maketitle
\vspace{-0.5cm}
\centering
    %\hspace*{0.5cm}
    \includegraphics[width=0.95\textwidth]{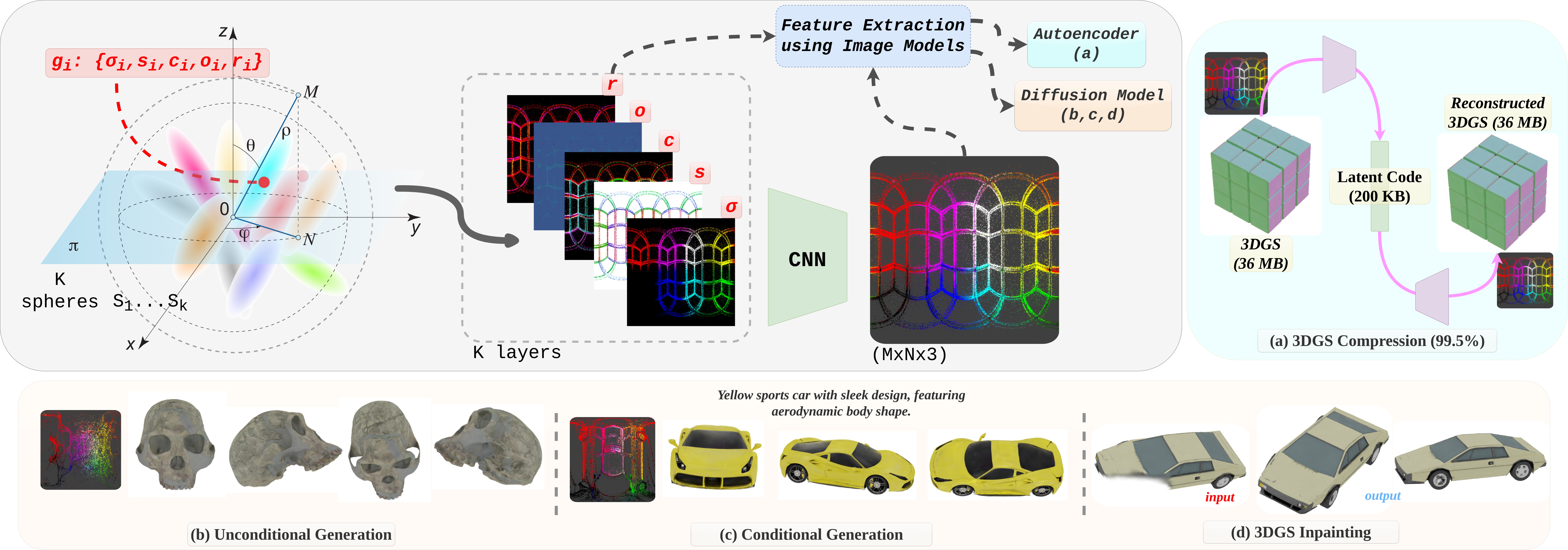}
    % \vspace*{-0.1cm}
    \captionof{figure}{\textit{
    We propose UVGS - an structured image-like representation for 3DGS obtained by spherical mapping of its primitives. The obtained UVGS maps can be further unified to a 3-channel 
    % ``3D-aware" 
    Super UVGS image to bridging the gap between 3DGS and existing image foundation models. We show Super UVGS can compress the 3DGS assets using pretrained image Autoencoders, and for direct unconditional and conditional 3DGS object generation using diffusion models. We also present one of the first experiments on 3DGS inpainting.
    % \srinath{This figure is a bit too busy, can be simplified, \eg~remove all notation.
    % Also, put high quality output results.}
    }}
    \vspace*{0.2cm}
    \label{fig:teaser}
}]

\begin{abstract}
\vspace*{-0.8cm}

% Following is the Dilin's suggested Abstract:

3D Gaussian Splatting (3DGS) has demonstrated superior quality in modeling 3D objects and scenes. However, generating 3DGS remains challenging due to their discrete, unstructured, and permutation-invariant nature. In this work, we present a simple yet effective method to overcome these challenges. We utilize spherical mapping to transform 3DGS into a structured 2D representation, termed UVGS. UVGS can be viewed as multi-channel images, with feature dimensions as a concatenation of Gaussian attributes such as position, scale, color, opacity, and rotation. We further find that these heterogeneous features can be compressed into a lower-dimensional (e.g., 3-channel) shared feature space using a carefully designed multi-branch network.
The compressed UVGS can be treated as typical RGB images. Remarkably, we discover that typical VAEs trained with latent diffusion models can directly generalize to this new representation without additional training. Our novel representation makes it effortless to leverage foundational 2D models, such as diffusion models, to directly model 3DGS. Additionally, one can simply increase the 2D UV resolution to accommodate more Gaussians, making UVGS a scalable solution compared to typical 3D backbones. This approach immediately unlocks various novel generation applications of 3DGS by inherently utilizing the already developed superior 2D generation capabilities. In our experiments, we demonstrate various unconditional, conditional generation, and inpainting applications of 3DGS based on diffusion models, which were previously non-trivial.

\end{abstract}    
\vspace{-0.8cm}
\section{Introduction}\label{sec:intro}
\vspace{-0.2cm}
%
% Contributions:
%
% 1. Dataset: from Objaverse mesh to 3DGS.\\
% - render 88 views\\
% - no Colmap
% 2. 3DGS to UVGS\\
% - spherical unwrapping\\
% - dynamic mapping using opacity \\
% - Compression using super UV gaussians. \\
% - address global correspondence
% 3. Mapping Networks\\
% - Multi-head networks\\
% - Pretrained Image Autoencoder
% 4. Caption Generation
% Why UV??
%
% - 3DGS is permutation invariant.\\
% - Although 3DGS has been widely studied in scene reconstruction tasks, its spatially unstructured nature presents a significant challenge when applied to mainstream generative modeling frameworks. \\
% - 3DGS is unstructured, thus poses a signifacnt challenge when applied to mainstream computer vision models, including AE, generative models. \\
% - A structured representation would eliminate the need of complex, specialized network designs that are often necessary with unstructured representations. \\
% - If we use a structured representation for 3DGS (spatial coherence), we can essentially use standard CNNs to capture the correlation among the neighboring gasussians, allowing efficient feature extraction. \\
% - Also address global correspondence.
%
% (First para: Highlight the need of 3D Object Reconstruction/Generation and the use of 3DGS.)
%

The creation of high-quality 3D content is essential in applications like virtual reality, game design, robotics, and  movie production, where realistic 3D representations play a critical role. Typical 3D representations like Neural Radiance Fields (NeRF)~\cite{nerf2021} are promising but require substantial computational resources, limiting their scalability for real-time applications. Moreover, NeRF is an implicit representation, which makes editing and manipulation challenging.
% \srinath{NeRF is also implicit making it hard to edit, etc.}
Recently, 3D Gaussian Splatting (3DGS) \cite{3dgs2023} emerged as a compelling alternative, enabling efficient and high-fidelity 3D rendering through a large set of Gaussian primitives that model spatial and visual properties.
As an explicit representation, 3DGS offers several advantages over NeRF.
% of easier manipulation and direct access to individual primitives. 
% \srinath{Explicit-ness of 3DGS is a big advantage.}
However, while 3DGS offers benefits in terms of speed and visual quality, its unstructured, permutation-invariant nature presents significant challenges for generative tasks. Much like point clouds, it lacks a coherent spatial structure, impeding its integration with conventional image-based generative models. This lack of structure and coherence among primitives hinders the application of image-based generative models~\cite{shapesplat2024, diffgs2024, gaussiancube2024}, which rely on structured data representations.
%and are unable to process unordered points effectively. 

Previous methods have tackled these challenges by transforming 3DGS into structured formats, such as voxel grids~\cite{gaussiancube2024, sdfusion2023, gvgen2025} or image-based representations like Splatter Image~\cite{splatterimage2024} or triplanes~\cite{zou2024triplane}. Other approaches employ diffusion models to directly predict 3DGS attributes~\cite{gsd2024}. These methods, while achieving impressive visual results, often require substantial computational resources, memory-intensive multi-view rendering, complex architectures limiting their scalability and flexibility for high-fidelity generation.
% Further, direct generation 
Generating and processing 3DGS directly by efficiently utilizing modern generative models like Variational Autoencoders (VAEs) and diffusion models is limited as the neural networks are not permutation invariant. % \srinath{incomplete sentence}

%Thus, an effective transformation of 3DGS into a structured representation is essential, enabling compatibility with image-based architectures and allowing for efficient local and global feature extraction without multi-view dependencies.

% \mj{motivating the problem here. May be need to add support with references.}
% To effectively leverage generative models for 3D Gaussian splatting, it is crucial to develop a structured representation for 3DGS that provides coherence without relying on multi-view rendering or other memory-intensive approaches. Such a transformation would allow compatibility with 2D image-based architectures, making it possible to efficiently extract local and global features from the 3D data.

% To this end, we propose the \emph{UV Gaussian Splatting (UVGS)}, a structured representation that leverages spherical mapping to organize the 3DGS attributes into a coherent 2D UV map. %By inscribing Gaussian splats within a spherical surface, UVGS transforms 3D Gaussian attributes—such as position, rotation, scale, opacity, and color—into an image-like 14-channel map, resolving the issue of permutation invariance.
% \mj{
To address these shortcomings, we introduce \textbf{UV Gaussian Splatting (UVGS)}, which provides a structured transformation of 3D Gaussian primitives into a 2D representation while preserving essential 3D information. 
We use spherical mapping~\cite{sphericalmapping2006} that inscribes Gaussian splats in a spherical surface, and projects attributes like position, rotation, scale, opacity, and color into an organized 14-channel image-like UV map.
This mapping introduces spatial structure, resolving issues of permutation invariance by introducing local correspondences between neighboring Gaussians and global coherence across the entire 3D object. 
%The result is a representation that is ``almost`` 3D \srinath{re-phrase as ``3D representation''}, but structured as a 2D map, enabling compatibility with powerful image-based neural network architectures. 
The result is a representation that functions as a ``3D representation" structured in a 2D map format, enabling compatibility with powerful image-based neural network architectures.

% }
% Our UVGS approach transforms 3DGS attributes, including position, rotation, scale, opacity, and color, into an organized 14-channel image-like representation that preserves both local and global correspondences. 
% This structured UVGS representation preserves both local correspondences among neighboring Gaussians and global coherence across the entire 3D object. 
% \dilin{i would suggest to explain local and global correspondence a little bit more as this is the main contribution of the paper..} 
% Thus, enabling the application of powerful image-based neural network architectures to unstructured 3D data, effectively bridging the gap between unstructured 3DGS and 2D image models. 
% \mj{Please add details and references as needed.} \ar{Need to add multi-layer unwrapping}. Additionally, we introduce multi-layer unwrapping in UVGS to enhance mapping fidelity across different scales.

% \dilin{highlight the magic of our method, it's "almost" a 3d representation, but it works as a 2d representation. zero-shot generalize to pretrained 2d foundation mdoels. while previous work, like triplanes, neural fields, occpuancy, voxels, point clouds, there first need a 3d backbones, it's not scalable and also requires 3d training data cannot benefit from the prior we learned from 2d easily.}

While UVGS introduces structure into 3D Gaussian Splatting, its full 14-channel attribute-specific representation presents challenges for direct integration with pretrained 2D generative models, as these models typically expect a simpler, image-compatible data. 
Each of its heterogeneous attributes—position, color, and transformation—has its own distinct distribution and resides in a separate feature space, making it challenging to represent the 3D object in a unified shared space.
%complicating efforts to collectively represent the 3D object.
 To address this, we introduce \textbf{Super UVGS}, a compact 3-channel representation that unifies these diverse attributes into a cohesive format. Using a carefully designed multi-branch mapping network, Super UVGS consolidates the distinct attribute spaces into a shared feature space, enabling a more collective representation of the object. This unified transformation not only facilitates zero-shot compatibility with pretrained 2D models but also optimizes memory usage and computational efficiency, making Super UVGS highly practical for large-scale 3D tasks.
 Unlike previous approaches that use Triplanes, voxels, occupancy grid, neural fields etc. and require specialized 3D architectures to train on 3D data, UVGS effortlessly leverages widely available pretrained 2D foundational models.
This zero-shot generalization capability allows UVGS to fully benefit from priors learned in 2D domains from large amount of data, improving both flexibility and scalability.
% Without adaptation, the 14-channel representation would require additional redesign or retraining of the 2D models.
%, limiting its zero-shot applicability. 
% To address this, we introduce \textbf{Super UVGS}, a compact 3-channel representation that transforms UVGS into a streamlined, image-compatible format. Using a multi-branch mapping network, Super UVGS efficiently consolidates the diverse Gaussian attributes into just three channels, allowing seamless integration with existing 2D models without compromising essential spatial information. This compression not only facilitates zero-shot use with 2D architectures but also reduces memory usage and computational demands, making Super UVGS highly practical for large-scale 3D tasks that require efficient processing.
% \dilin{missing the motivation on why we do want to introduce super uvgs, maybe move the second sentence to the first?} 
%Building on UVGS, we introduce \emph{Super UVGS}, a compact 3-channel representation achieved through a multi-branch mapping network. This network efficiently encodes distinct distributions of Gaussian attributes, producing a streamlined, image-compatible format without quality loss. This compact representation allows us to directly apply image-based generative models, such as auto-encoders and diffusion models, \mj{without additional training or architectural modifications}, facilitating conditional generation, in-painting, and  high-quality 3D asset synthesis. \dilin{highlight we don't even need training?}
%
To sum up our main contributions are:
\begin{itemize}
    \item \emph{Efficient Structured Representation of 3DGS}: We present UVGS, an image-like representation that solves permutation invariance and unstructured nature of discrete 3DGS through spherical mapping, making direct feature extraction possible by organizing unordered points into a coherent 2D representation compatible with 2D models.
    % \srinath{What is ``efficient'' about this representation?}
    \item \emph{Compact and Scalable Super UVGS Representation}: To address scalability while dealing with large scale 3DGS points and enabling the direct integration of pre-trained 2D foundation models, we introduce Super UVGS - a low-dimensional version of UVGS maps that retains high fidelity features while reducing memory overhead. 
    % Using a multi-branch mapping network, Super UVGS enables large-scale 3D tasks with minimal computational overhead. 
    %This compact format maintains high fidelity while reducing memory usage (\mj{refer to an experiment}), enhancing scalability for complex 3D tasks.
    % \srinath{briefly say why we need super uvgs}
    % \item \emph{Direct Application of Image-Based Models for 3D Generation}: By transforming 3DGS into a structured image-like representation, we enable the use of pre-trained image-based models without specialized 3D architectures, achieving multiview consistency and improved generation quality.
    \item \emph{Diverse 3D Applications}: Our approach unlocks seamless integration of 3DGS with pre-trained 2D foundation models for various tasks, including unconditional and conditional generation of 3DGS. 
    % Direct integration with 2D diffusion models marks one of the first instances of applying image-based generative models to 3D Gaussian data.
    %We validate our approach with various applications, including conditional generation, unconditional generation, and 3D inpainting using diffusion models. This marks one of the first experiments on applying image-based generative models directly to 3D Gaussian data.
%
\end{itemize}

\vspace{-0.1cm}
\section{Related Work}\label{sec:literature}
% \ar{in progress... please add anything that is relevant}
% \dilin{UV mapping..} \ap{} \ar{Do we really need that?}

\begin{figure*}[t]
\centering
\includegraphics[width=0.92\linewidth]{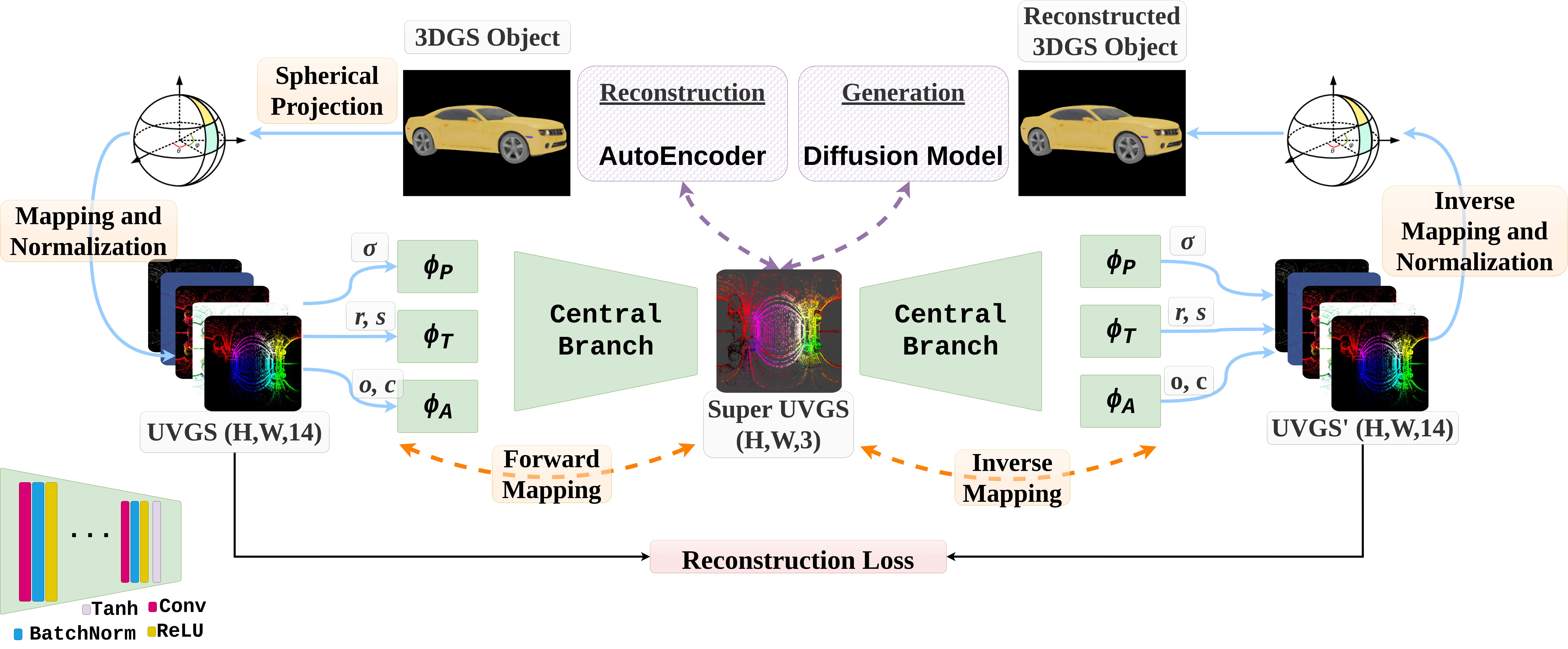} 
\vspace{-0.2cm}
\caption{The input 3DGS object is first converted to UVGS maps through spherical mapping. 
We use a multibranch forward mapping network to convert the obtained 14-channel UVGS to a compact 3-channel Super UVGS image. This represents the 3DGS object in a structured manner and can be used with image foundation models for reconstruction or generation. 
The Super UVGS is mapped back to UVGS through branched inverse mapping, which in turn can be reconstructed back to the 3DGS object through inverse spherical mapping. 
% The forward and inverse mapping networks are trained together using reconstruction losses. 
}
\label{fig:architecture}
\vspace{-0.5cm}
\end{figure*}

\noindent\textbf{3D Generative/Reconstruction Models for Objects}:
% \ar{Discuss about various 3D representations and 3D asset generation/reconstruction. Introduce 3DGS and recent advancements in object representation.}
%
Generation or reconstruction of 3D assets has been a long standing task \cite{dreamfusion2022, magic3d2023, prolificdreamer2024, dreambooth3d2023, latentnerf2023, gaussiandreamer2023, dreamcraft3d2023, makeit3d2023, eg3d2022, albedogan2024, 3dfacecam2023, tsdf2024, genheld2024}. Previous reconstruction approaches like NeRF~\cite{nerf2021, mipnerf2021} are often slow and do not provide a defining geometry \cite{latentnerf2023, dreambooth3d2023, neaf2023, udiff2024, tensorf2022, dnerf2021, plenoxels2022, neus22023, neusurf2024, geodream2023, zeroshot2024}. Advancements in the field led to the emergence of explicit voxel grid based representations that encode colors and opacities directly \cite{sdfusion2023, diffrf2023, volumediffusion2023}. These approaches achieve significant speed ups compared to the NeRF based approaches, but they can't produce high fidelity assets due to the low resolution of voxel grids. On the other hand, triplane representation \cite{3dgen2023, triplanemeetsgs2024, triplanediff2023, rodin2023} provides a trade-off between the quality and memory utilization. 
Another line of work \cite{geometry_img_diffusion2024, omages2024} splits the input mesh into different patches and simplifies the object generation problem to an image generation problem. 
However such methods rely on either cutting through the mesh to create a geometry image~\cite{geometry_img_diffusion2024} or rely on an existing UV representation of the geometry and utilize subset of existing UV islands~\cite{omages2024} resulting in loss of details. 
% However, the fact they assumes meshes to have patch decomposition makes them constrained and also results in visible cracks~\cite{geometry_img_diffusion2024} and artifacts~\cite{omages2024} in synthesized objects.
Recently, there has been a notable advancement in 3D Gaussian Splatting (3DGS) for the representation of objects and scenes leding to the emergence of 3DGS showcasing impressive real-time results in reconstruction and generation tasks \cite{3dgs2023, 4dgs_realtimescene2024, dngaussian2024, gaussiandreamer2023, dreamgaussian2023, viewconsistentediting3dgs2025, gsedit2024, grm2024, triplanemeetsgs2024, gslrm2025, gvgen2025}. 
Recent advances in the 3D generative models for asset synthesis using the existing geometries like NeRF, voxel grids, or triplane geometries \cite{dreamgaussian2023, grm2024, triplanemeetsgs2024, gslrm2025, get3d2022, difftf2023} leverage generative models \cite{stable_diffusion, ddim} and the existing 3D datasets~\cite{shapenet2015, objaverse2023}. However, most of the works employing 3DGS or other representations use multiview rendering and Score Distillation Sampling (SDS) to achieve convincing generation and reconstruction capabilities \cite{textto3Dusinggs2024, dreamfusion2022}. These approaches demand high memory and compute resources and are often quite slow in optimizing due to per scene optimization.

%Direct Learning on Trained 3DGS / 
\noindent\textbf{Giving Structures to Discrete Gaussians}:
% \dilin{there are some repeating sentences here, eg. SDS has been mentioned before. Also we do we want to discuss SDS, SDS is an optimization technique, while we focus mostly presentation? these two are orthogonal? this could be even one of our followups, building on UVGS and using SDS to optimize very high quality 3DGS for data curation.}
Although, 3DGS has led to breakthrough in the reconstruction field by demonstrating superior performance in multiple domains, the generation of 3DGS directly remains challenging due to its discreteness and unstructured nature~\cite{diffgs2024, gaussiancube2024}. 
% The 3DGS framework is discrete and permutation invariant, meaning that randomly shuffling the order of points does not impact the resultant shape of the object. Consequently, it can be treated as a set of Gaussian points with no correspondence between them. Additionally, these points are unstructured in the spatial domain. 
These characteristics present substantial challenges when integrating them with conventional computer vision models, like Autoencoders and generative models \cite{diffgs2024}.
The research in direct learning of trained 3DGS primitives is largely unexplored~\cite{shapesplat2024}. 
% Most existing generation and reconstruction methods that utilize 3DGS employ multi-view rendering and SDS loss \cite{dreamgaussian2023, gaussiandreamer2023, dngaussian2024, gvgen2025, gslrm2025} to generate various views and optimize the Gaussians. 
% This approach is both time-consuming and memory-inefficient, necessitating complex architectures to generate 3DGS assets. 
%
Some efforts attempt to address this by directly predicting 3DGS attributes using diffusion models~\cite{gsd2024} while others
% These are constrained to generating a limited number of 3DGS Gaussian points due to the unstructured and permutation-invariant nature of 3DGS.
like Splatter Image~\cite{splatterimage2024} project Gaussian objects into image-based representations through direct 3D-unaware projection. 
These methods struggle with maintaining multiview consistency, as the model only infers seen poses correctly, while hallucinating for unseen poses. 
% The scarcity of extensive 3D data hampers the ability to produce high-quality results for such methods. 
%

Concurrent works~\cite{gaussiancube2024, gvgen2025} follow the voxel-based representations to transport Gaussians into structural voxel grids with volume generation models for generating Gaussians. However, these methods are computationally expensive for high-resolution voxels, face difficulties in preserving high-quality Gaussian reconstructions due to information loss during voxelization. 
% , and a restricted number of generated Gaussians constrained by voxel resolutions. 
% For example, GaussianCube~\cite{gaussiancube2024} can accommodating only up to 32,000 Gaussians. However, in practice, representing an object adequately with 3DGS may require up to a few hundred thousand Gaussians~\cite{3dgs2023, diffgs2024}.
% A very recent concurrent work, 
DiffGS~\cite{diffgs2024} tries to solve the above issues by proposing three continuous functions to represent 3DGS. However, it is limited to only category-level generation and learning generic probability functions for all the categories poses significant compute and design challenges.
In contrary, we introduce an efficient way to give structures to discrete Gaussians by taking inspiration from the developments in 3D graphics. 
Our method does not require any learning to map an unstructured set of Gaussians to this efficient and structured representation (termed UVGS). 
The proposed representation provides local and global correspondence among different Gaussian points making the widely available existing computer vision frameworks learn and extract underlying features from them.
\vspace{-0.15cm}
\section{Methodology}

\noindent\textbf{Preliminaries}:
% \dilin{just a minor suggestion, not sure if you would like to use $\mu$ to represent the center, $\sigma$ is often used as the standard derivation.} \ar{I was planning to use $\mu$ for VAE, but we didn't end up describing it. I'll change if we have time at the end.}
3DGS represents an object or a scene with a collection of Gaussians primitives to model the geometry and view-dependent appearance. 
% \dilin{we don't have view-dependent effect?} \ar{Nope}. 
For a 3DGS set, \(G =\{g_i\}_{i=1}^N\), representing an object with $N$ individual Gaussians, the geometry of the $i^{th}$ Gaussian is explicitly parameterized via 3D covariance matrix $\Sigma_i$ and it's center $\sigma_i \in \mathbb{R}^3$ as:
% \ns{put gaussian equation}
\setlength{\abovedisplayskip}{3pt}
\setlength{\belowdisplayskip}{3pt}
$
    g_i(x) = e^{(-\frac{1}{2} (x - \sigma_i)^T \Sigma^{-1} (x - \sigma_i))}
$
where, the covariance matrix $\Sigma_i = r_i s_i s_i^T r_i^T$ is factorized into a rotation matrix $r_i \in \mathbb{R}^4$ and a scale matrix $s_i \in \mathbb{R}^3$. 
The appearance of the $i-th$ Gaussian is represented by a color value $c_i \in \mathbb{R}^3$ and an opacity value $o_i \in R$. 
In practice, the color is represented by a series of Spherical Harmonics (SH) coefficients, but for simplicity, we represent the view-independent color by just RGB values. 
% \ns{In practice I'm not sure if we can say that RGB values are used since several papers use SH directly}
% \ar{I see in most of the generative and reconstruction papers, people tend to ignore SH.}
Thus, a single Gaussian can be represented by a set of five attributes as $ g_i = \{ \sigma_i, ~r_i, ~s_i, ~o_i, ~c_i \} \in \mathbb{R}^{14}$, and the entire 3DGS can be represented by a set of $N$ such Gaussians as:
$G = \{ \{ \sigma_i, ~r_i, ~s_i, ~o_i, ~c_i \} \}_{i=1}^N$.

\subsection{Spherical Mapping}
3DGS is represented as a permutation invariant set with no structural correspondence among different Gaussians $g_i$, making it challenging to extract meaningful features from this set containing a few hundred thousands of them using neural networks. 
% Thus, it is fairly complicated and non-trivial to use the existing large number of neural network frameworks such as, generative models, autoencoders, or anything in general for feature extraction from
% \ns{same with above: SUCH AS ???} 
% \ar{Such as, generative models, autoencoders, anything in general for feature extraction.} 
% this representation. \dilin{no need to repeat (first sentence)?}
% \ar{Basically, what I want to emphasize that due to the unstructured and permutation invariant nature of 3DGS, we can not use the existing models directly to process 3DGS and extract local and global features. The same motivation is mentioned in the recently released DiffGS and other similar papers. Shall I cite them here??}
% \ns{maybe let's clarify a little what we mean by this sentence and provide a stronger motivation for the next sentence}
To address this, we introduce a novel representation that gives structure to this unstructured set of points and solves the permutation invariance issue for faster and better feature extraction. We propose to accomplish this by employing spherical mapping to map the 3DGS primitives to an image-like representation that is both invariant to random shuffling of 3DGS points and well structured. % Nikos: Minor comment but we haven't explained the novel aspect of it just yet. I'll eventually rephrase this
% \srinath{why spherical mapping? Why not cylindrical or cubic?}
% We prefer spherical mapping as others generally fails to capture the top and bottom parts of the object in the same UV map, and can introduce distortions for objects that extend far in the Z-direction.

We begin the mapping 
% that is described in detail in the supplementary 
% \dilin{cannot be in supplementary :) ?}, 
by inscribing the 3DGS object into a sphere with the same center as the object in the canonical space.
% \ns{inscribing??? can we maybe explain this with a few words}
Inscribing a 3DGS object into a sphere involves enclosing the object within a sphere. This begins by determining the geometric center of the object. The next step is to calculate the radius of the sphere, which is achieved by measuring the Euclidean distance from the center to the farthest point on the object. The radius of the sphere is defined such that the sphere fully encloses the object. The sphere acts as a bounding volume for the entire object.

We consider each Gaussian $g_i$ in 3D to be centered at the mean position represented by $\sigma_i$ with Cartesian coordinates $(x_i,y_i,z_i)$. 
The aim is to get the spherical coordinates $(\rho_i,\theta_i,\phi_i)$ for each Gaussian $g_i$. To do so, we calculate the azimuthal $\theta_i$ and polar $\phi_i$ angles for each $g_i$ along with the distance from the origin to the point, $\rho_i$. The spherical radius is defined as \(\rho_i = \sqrt{x_i^2 + y_i^2 + z_i^2}\), the azimuthal angle as \( \theta_i = \tan^{-1}(y_i, x_i)\), while the  polar angle as \( \phi_i = \cos^{-1}(z_i, \rho_i) \). 
The azimuthal and polar angles are then normalized, such that we can map them on a 2D UV map of $M \times N$ dimensionality with 14-channels. 
% \ns{Please Merge the following to one equation for theta and phi as converting to angles is not really interesting to take that much space :) }
\(\theta_i\) and \(\phi_i\) are converted to degrees and mapped to UV image coordinates: \(\theta_{i~\text{scaled}} = \left\lfloor \frac{\pi + \theta_{i}}{2\pi} \times \text{M} \right\rfloor, \; \; \phi_{i~\text{scaled}} = \left\lfloor \frac{\phi_i}{\pi} \times \text{N} \right\rfloor\)
   % \[
   % \theta_{i~\text{scaled}} = \left\lfloor \frac{\pi + \theta_{i}}{2\pi} \times \text{M} \right\rfloor
   % \]
   % \[
   % \phi_{i~\text{scaled}} = \left\lfloor \frac{\phi_i}{\pi} \times \text{N} \right\rfloor
   % \]
Each channel in the UV map stores 3DGS attributes, including $\{ \sigma_i, ~r_i, ~s_i, ~o_i, ~c_i \} \in \mathbb{R}^{14}$. We refer this 14-channel UV map as \emph{UVGS}, $U \in \mathbb{R}^{M\times N \times14}$ defined as:
\begin{equation}
    \text{U}[\phi_{i~\text{scaled}}, \theta_{i~\text{scaled}}, :] = [ \sigma_i, ~r_i, ~s_i, ~o_i, ~c_i ]. %\nonumber
\end{equation}

This transformed UVGS representation provides spatial coherence and solves the permutation invariance problem as any random arrangement of points will now map to the same UVGS representation $U$.
It should be noted that this kind of transformation will also preserve the spatial correlation between the Gaussian points in 3D and transform them to 2D UV maps by mapping them to neighboring pixels. 
This provides both the local level correspondence among the neighboring Gaussians and the overall global correspondence for the object. 
Thus, solving the unstructured and discreteness problems.
This enables standard neural network architectures (\eg CNNs) to effectively capture correlations among neighboring Gaussians for efficient feature extraction. 
% \dilin{even though it might be obvious already, but i feel like it would benefit to further discuss how nearby gaussians would be mapped to nearby pixels in the UVGS representation.}
% \ar{answered above}
% \st{In our case, we fix the size of the UV maps to $512\times512$. Through our experiments, we realized that UV maps of size $512 \times 512$ are sufficient to represent objects in our dataset and capable of storing upto $262K$ Gaussians.}\ar{transferred to experiments}

% \st{However, the use of this UVGS representation with the existing image based models still remains a challenge as most of these models are structurally defined to operate on a 3-channel image.} 
% \ar{To further unify the extracted position ($\sigma$), transformation ($r, s$), and color ($c, o$) maps in a same feature space and use the existing image foundation models, we map...}
% \dilin{This seems like a very strange argument to me. I guess, the point to be able to directly reuse all pretrained foundation models? since that you want to map the channel to 3? But there's no constraints on the network design? Also wouldn't it make more sense to say that because we now have scales and colors in the channel dimension, which have very different meanings, we would like to further unify them in the same feature space?} 
% \ar{Yes, I think this makes more sense and also provides a sufficient explanation to why we had to use branching.}
%
% \ns{In the previous sentece explain more clearly what do you mean by "application". I know it'll be clear later but currently it's not. }
To further unify the extracted position ($\sigma$), transformation ($r, s$), and color ($c, o$) maps in a same feature space and use the existing image foundation models, we map the obtained UVGS $U \in \mathbb{R}^{M\times N \times14}$ further to a 3-channel image $S \in \mathbb{R}^{M\times N \times 3}$ (termed as Super UVGS), using a Convolutional Neural Network (CNN).
% We might get a Q why we do that this way vs alternatives. Let's discuss a little 
A multi-branch forward mapping network is employed to map $U \in \mathbb{R}^{M\times N \times14}$ to the 3-channel Super UVGS $S \in \mathbb{R}^{M\times N \times 3}$. We provide all technical details in Sec.~\ref{ssec:mapping} and the supplementary material. 
% on that in Section~\ref{ssec:mapping}
% \ns{Can we draw a parallelism here with neural textures that embed in HxWxC dims a feature map that maps to textures while keep the first 3 channels for rgb.}
% \ar{"kind of" for UVGS. unlike neural textures, we also store position and transformation information in the maps.}

The Super UVGS representation effectively retains all the details of 3DGS attributes and can be directly utilized with existing widely available image-based models. 
We demonstrate this by showing perfect reconstruction of 3DGS object from Super UVGS image in the experiments section.
% \ns{where do we demonstrate this? Let's either get into details here or tell them that we do it in the experimental section} \ar{We are demonstrating this by showing almost perfect reconstruction of 3DGS object from Super UVGS image.}
This semantically structured representation $S$ offers both local and global correspondence 
in representing Gaussian attributes and needs relatively less storage. 
% \dilin{local and global correspondence was mentioned a couple of times but without an explicit explaintion, as well as why it's important} 
% \ar{answered above in RED}
% To reconstruct the object back from a Super UVGS represent, we train an inverse mapping network that maps a 3-channel Super UVGS image back to a 14-channel UV map representing all five 3DGS attributes. 
% Through inverse spherical projection, we can easily reconstruct the 3DGS object with minimal computation. 

\begin{figure}[t]
\centering
\includegraphics[width=0.85\columnwidth]{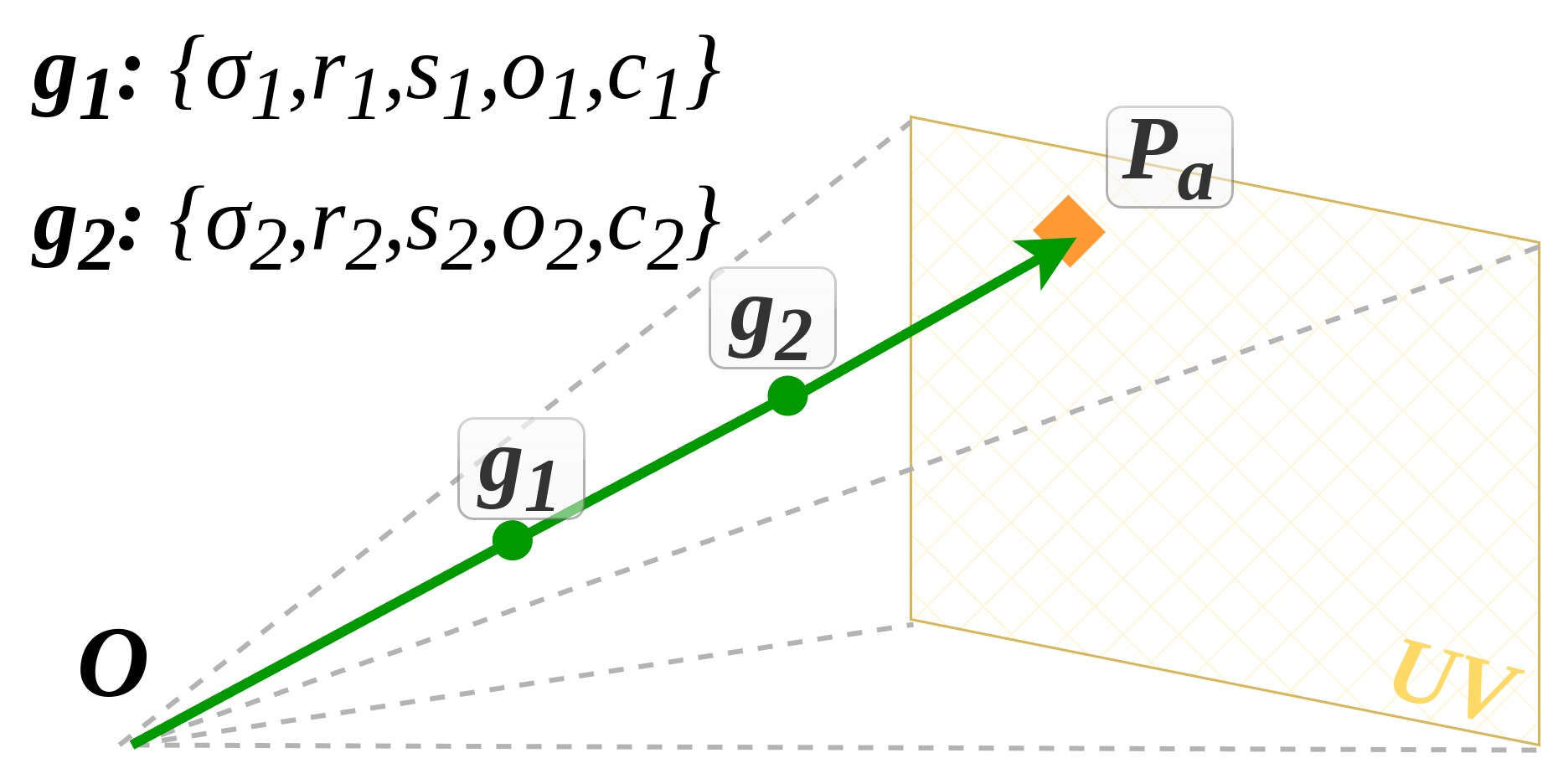} 
\vspace{-0.2cm}
\caption{
\textbf{Dynamic Selection}. In spherical mapping of 3DGS points to UV maps, multiple points may map to the same pixel, creating a many-to-one issue. Our Dynamic Selection approach addresses this by retaining the attributes of the point with the highest opacity per pixel on the same ray.
% \dilin{should add an illustration of the multi-layer here, otherwise, the figure alone only delivers a negative information.}
}
\label{fig:dynamic_sel}
\vspace{-0.4cm}
\end{figure}
\vspace{-0.2cm}

% \vspace{-0.1cm}
\subsection{Dynamic GS Selection and Multiple Layers}
\begin{figure}[t]
\centering
\includegraphics[width=3.2in]{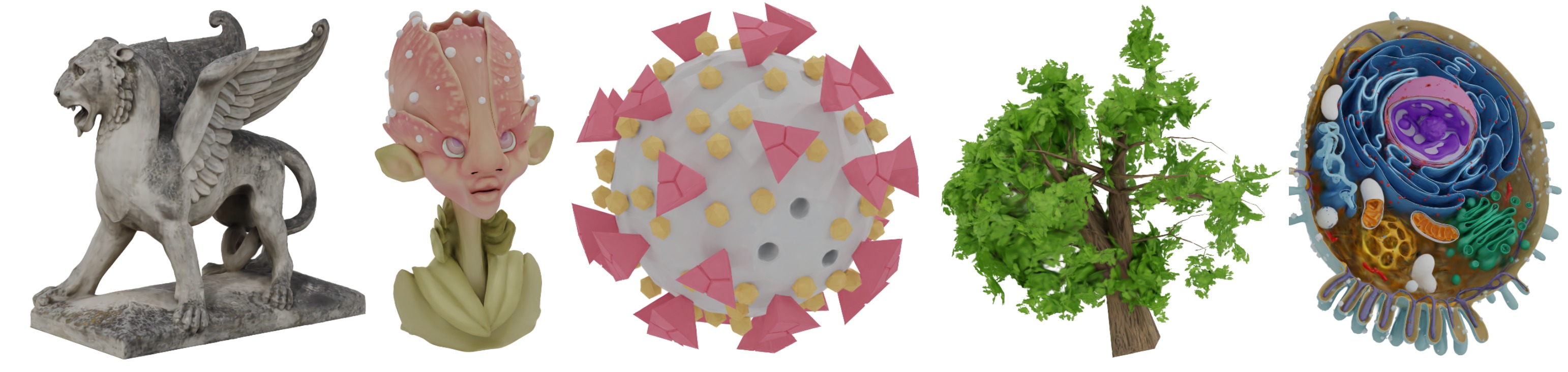} 
\vspace{-0.3cm}
\caption{
Complex object reconstructions (K=4) using pretrained image-based autoencoder.
}
\vspace{-0.6cm}
\label{fig:complex_recons}
\end{figure}

% \dilin{i would suggest to move this section right after sec 3.1} \ar{Moved here from 3.4}
When projecting 3DGS points to UV maps using spherical mapping, multiple points may map to the same pixel in UV space as shown in Fig.~\ref{fig:dynamic_sel}. 
Two 3DGS points $( g_1 )$ and $( g_2 )$ map to the same pixel on UV map $( P_a )$ causing many-to-one mapping issue. 
% \ns{Re-iterate here why we don't want that which will then motivate the next sentence}
To address this, we propose a Dynamic Selection approach where each UV pixel retains the 3DGS attribute with the highest opacity intersecting the same ray.
Using the same example in Fig.~\ref{fig:dynamic_sel}, if opacity $o_1$ of Gaussian $g_1$ is less than opacity $o_2$ of $g_2$. Then only $g_2$ will be stored in the UV map at pixel $P_a$. 
We observed that this method helps maintain the geometry and appearance of the 3DGS object while resolving many-to-one mapping issues with minimal quality loss.
% \st{Recent literature has also shown that a single Gaussian point can represent both simple and somewhat complex geometries with sufficient quality.} \dilin{this is kind of confusing, because we're discussing a multi-layer solution right afterwards}. % CITE Splatter-Image and LGM papers and whatever else
%
For more complex objects or real-world scene representation, we stack multiple such layers of UV maps, where each UVGS pixel now holds attributes of the top-K opacity values of 3DGS primitives. 
This can be accomplished by inscribing the 3DGS object inside multiple spheres where each sphere maps the 3DGS attribute corresponding to the top-$K^{th}$ opacity value along the same ray. More details on this are presented in the supplementary.
To show the effectiveness of proposed UVGS maps in capturing the intricacies of a complex objects, we used a pretrained image based autoencoder to reconstruct objects using a 4 layer UVGS as shown in Fig.~\ref{fig:complex_recons}. 
% \ar{Do we need to put a reference to Supplementary for scenes or something?}
% \ns{understand should be accomplished maybe???}\ar{fair point!}
% \ar{Moved the following to experiments section:}
% \st{Table XX compares 1-8 layer UV maps. 
% Fig.~[XX] in supplementary shows that this multi-layer UVGS mapping can be used to even map real-world complex scenes to structured UV maps and can be reconstructed back with high-fidelity. 
% Interestingly, for most of the objects in our dataset with convex geometry, a single-layer UVGS map (K=1) suffices without significant quality loss. Therefore, we assume a single-layer UV map for our experiments, unless stated otherwise. }
% \ns{lets not use concerned anywhere in the paper :) you can write general purpose objects, objects in our dataset or something more descriptive}
% \dilin{i feel the following is not necessary..it's very clear before already.} \st{Note that a multi-layer UVGS map should not be confused with a 14-channel UVGS $U \in \mathbb{R}^{M\times N \times14}$ representing the 5 attributes. A 2-layer UVGS map will have \(14 \times 2 = 28\) channels, with each attribute represented by 2 identical attribute specific layers. 
% Similarly, a 4-layer UVGS map will have \(14 \times 4 = 56\) channels.}

\vspace{-0.1cm}
\subsection{Mapping Networks}\label{ssec:mapping}
% FORWARD MAPPING
Our goal is to bring the extracted UVGS maps to a common feature space to better represent the object collectively and to make the 14-channel UVGS representations $U \in \mathbb{R}^{M\times N \times14}$ work with the widely available image based foundation models. To accomplish this, we map it to a 3-channel image which can be easily processed by the existing architectures while also maintaining the spatial correspondence. 
%
% \ar{To bring the extracted 3DGS attribute maps to a common feature space to better represent the object collectively and to make...}
% to make the 14-channel UVGS representations $U \in \mathbb{R}^{M\times N \times14}$ work with the widely available image based foundation models 
% \ar{, we map it to a 3-channel image which can be easily processed by the existing architectures while also maintaining the spatial correspondence. }
% \dilin{without the need of any type of training or finetuning?} \ar{Yes, in case of AE and VAE.}
% \ns{Is it really imperative? What other options exist and why we didn't go down these paths}\ar{todo}
%
We design a simple yet effective multi-branch CNN to extract features from different UVGS attributes and map them to a 3-channel feature-rich image, termed Super UVGS. The structured UVGS maps provides local and global features that can be learned by a CNN.
% \dilin{need to explain the intuition a little bit; otherwise, the setup doesn't look correctly. Essentially, there is redundancy between gaussians, and a CNN can encode some of this information implicitly. That's why we can compress the data.}\ar{answered above} \ap{this is still not answered. The intuition is redundancy reduction which is implicitly represented by CNN}

\noindent\textbf{Forward Mapping}
The first layer is a set of three mapping branches for position, transform, and appearance ($\phi^f_{P}, ~\phi^f_{T}, ~\phi^f_{A}$) respectively. We refer to them as position, transformation, and appearance branch. 
The position branch takes the mean position ($\sigma$) as an input and processes it to give a position feature map $M_{P}$.
Similarly, the transformation branch takes the rotation ($r$) and scale ($s$) together to generate another feature map $M_{T}$.
The last, appearance branch takes the color ($c$) and opacity ($o$) together to produce another feature map $M_{A}$. 
All the three features maps from position, transformation, and appearance branch are concatenated to get a final feature map, before passing them to the next module, called the Central Branch. The central branch ($\phi^f_{C}$) is composed of multiple hidden Convolution layers, where each layer is followed by BatchNorm and ReLu activation.
The last layer of the central branch is activated using $tanh$ to ensure the Super UVGS does not take any ambiguous value resulting in gradient explosion or undesired artifacts. 
The obtained Super UVGS $S$ representation squeezes all the 3DGS attributes to a 3 dimensional image while also maintaining local and global structural correspondence among them. 
% The intuiting behind branching is explained in the end of this section. \dilin{why the end?} \ar{To not discontinue the methodology flow. I can put it here if that makes more sense.}
% \ns{Please discuss here in detail the intuition behind this architectural design  mentioned above and what benefits we get from the separate branches etc.} \ar{I've put this in the end of this section. Would to suggest moving it here?}
%

% REVERSE MAPPING
\noindent\textbf{Inverse Mapping}
We design an inverse mapping network that aims to map the obtained 3-channel Super UVGS image $S \in \mathbb{R}^{M\times N \times 3}$ back to the UVGS maps to obtain each of the five different 3DGS attributes $\{ \sigma, ~r, ~s, ~o, ~c \}$. 
% \ns{Why do we need that. Atm is not motivated... you do it further down but you need to start with it}
The inverse mapping network simply follows the forward mapping network architecture in the reverse order, where at first, we put the Central Branch ($\phi_{iC}$) followed by attribute specific position, transformation, and appearance branches ($\phi_{iP}, ~\phi_{iT}, ~\phi_{iA}$). We provide more details on mapping network in the supplementary.

% RATIONALE BEHING BRANCHING
\noindent\textbf{Branched mapping layers}: 
% \dilin{this paragraph seems quite long..} \ar{made it short}
The rationale behind using branched mapping layers in both forward and reverse mapping networks is to prevent the incompatibility issues arising due the the different value distribution of 3DGS attributes. 
Note that the disparate distributions of values within each set of attributes in 3D Gaussian Splatting, (\ie, mean position, transformation, and color), pose a challenge to the model when processed collectively. 
For instance, neighboring Gaussians in UVGS maps show smooth changes in position and color values but typically have large variations in rotation, scale, and opacity values. This results in gradient anomalies and slow convergence.
To address this, we propose a multi-branch network architecture, where attribute-specific branches implicitly learn to process these distinct attribute specific properties, focusing on their unique features before passing them to the central branch. The central branch receives a concatenated stack of processed attributes and exploits the correlation between them by extracting local feature correspondences. This information is then mapped to a 3-channel Super UVGS image, effectively capturing the complex relationships between the various attributes.
 This approach enables our network to manage diverse attribute distributions, resulting in faster convergence, improved accuracy, and specialized processing for each attribute set.

\noindent\textbf{Reconstruction Losses}:
Since the obtained UVGS maps have both local and global features, we opted for image-based losses to train the overall architecture. 
We use a set of Mean Squared Error (MSE) and Learned Perceptual Image Patch Similarity (LPIPS)~\cite{zhang2018perceptual}  between the obtained UVGS from spherical mapping $U$ and the predicted UVGS $\hat{U}$ from inverse mapping network. 
% \dilin{clarifying LPIPS is only applied to colors?}. \ar{No, for 4 attributes} 
We calculate the LPIPS loss over four attributes of 3DGS including mean position ($\sigma$), view independent color ($c$), scale ($s$), and rotation ($r$). 
The overall LPIPS loss for UV maps can be written as a linear sum of individual attribute loss terms as: 
% \dilin{i thought lpips is only pretrained on image space?} \ar{Yes, that is true. But similar to other models, even LPIPS can be effectively applied on individual imge-like UV maps to perform feature matching.}
\begin{equation}
    \mathcal{L}_{UV-lpips} =\mathcal{L}_{\sigma} + \mathcal{L}_{s} + \mathcal{L}_{r} + \mathcal{L}_{c}
\end{equation}
The overall loss function for the training can be written as: \(\mathcal{L}_{uvgs} = \mathcal{L}_{mse} + \lambda . \mathcal{L}_{UV-lpips}\) where $\lambda$ is a scalar and varied from $0$ to $10$ during the course of training.
% \ns{you have 2 lpips losses in equations being equal to different things. While I know what you mean this is not clear enough so I'd suggest we rephrase or add different indices in the losses}

% \subsection{UVGS AutoEncoder \& Latent Diffusion Model}
% \dilin{maybe treat 3.4 as a new section, like Applications?} \ar{Trying to move this section to Experiments entirely - WIP.}

% % \ns{This needs to be motivated in a more clear manner and make a strong case. Let's provide details why are we doing this}

\vspace{-0.2cm}
\section{Experiments}
\vspace{-0.1cm}

\begin{figure*}[t]
\centering
\includegraphics[width=0.98\linewidth]{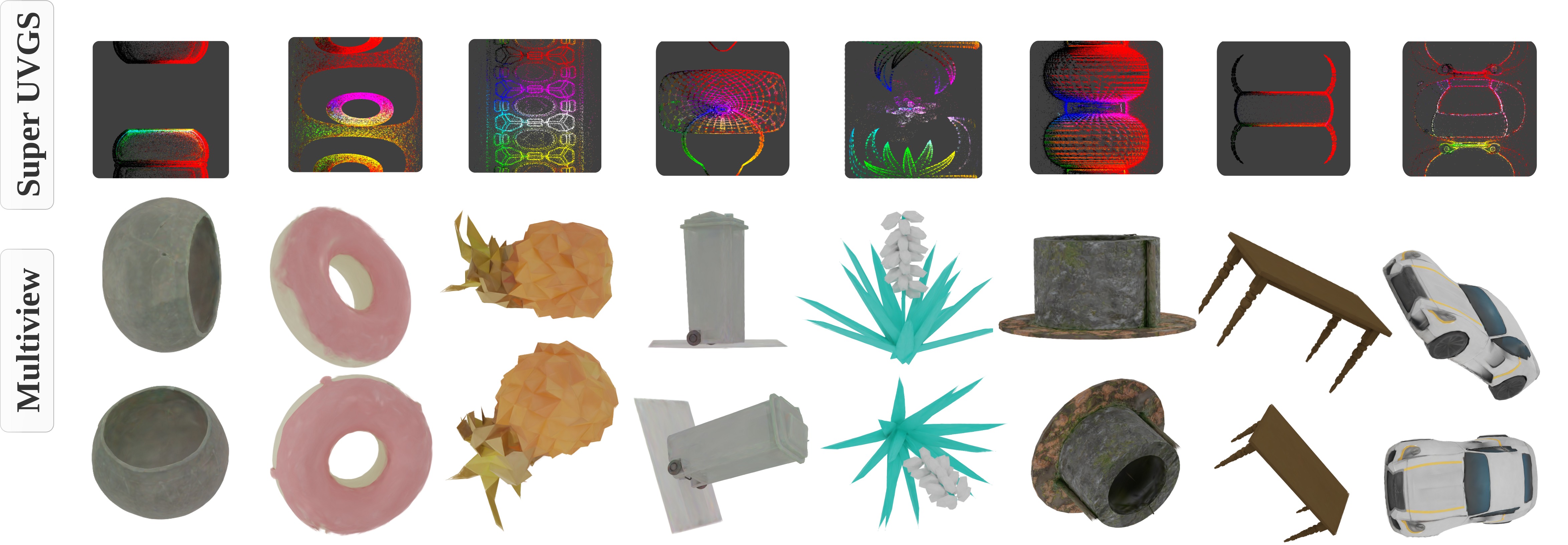} 
\vspace{-0.2cm}
\caption{
Figure shows a wide variety of high-quality unconditional generation result from our method using Latent Diffusion Model. We train an LDM to randomly sample Super UVGS images from random noise. The Super UVGS can be converted to 3DGS object using inverse mapping network and inverse spherical projection. The unconditional generation model was trained on Objaverse dataset.
}
\label{fig:unconditional}
\vspace{-0.4cm}
\end{figure*}

\noindent\textbf{3DGS Dataset and UV maps}
To train the mapping networks and learn a latent space for unconditional and conditional sampling, we need large amount of 3DGS assets. 
However, there's a lack of such a large-scale dataset for high quality 3DGS assets. To this end, we create a custom large scale dataset by converting the Objaverse~\cite{objaverse2023} meshes into 3DGS representation \footnote{Sketchfab data was filtered out of training data due to its license}.
% \ns{\textbf{TO DO}: Dilin/Aayush let's find out the best way to discuss this to not raise eyebrows}
We start by designing a scene of 88 cameras in a canonical space and use it to capture Objaverse objects from various angles covering all of the object views.
The 88 rendered views from different angles are then used to train a 3DGS for 10K iterations using \cite{3dgs2023}. 
This way, we create a high-quality and large-scale 3DGS dataset of $\mytilde 400K$ objects and scenes from Objaverse.
We only use static scenes or objects from Objaverse.
% \dilin{not sure why do we want to mention colmap here.}
% \st{Interestingly, we did not use COLMAP to get the extrinsic camera parameters, but instead we directly converted the Blender scene cameras into COLMAP format to use them for 3DGS fitting, thus avoiding loose feature matching arising due to COLMAP on simpler objects or scenes.}
After fitting all the object to 3DGS representation, we convert the objects to the corresponding UV maps (\ie, UVGS) through Spherical Mapping as illustrated in Fig.~\ref{fig:architecture}. 
For the course of our experiments, we only map the objects to a single layer UV maps as it was sufficient to represent the general purpose Objaverse objects with minimal quality loss. 
Through mapping, we gathered a UVGS dataset of  $\mytilde 400K$ maps.
We fix the size of the UVGS maps to $512\times512$. Through our experiments, we found that UV maps of size $512 \times 512$ are sufficient to represent objects in our dataset and capable of storing upto $262K$ unique Gaussians. 
Table~\ref{table:psnr} compares 1-4 layer UV maps. 
We also did experiments on ShapeNet~\cite{shapenet2015} cars dataset for evaluation purposes.
% \ar{\textbf{Do we need the following line in paper?} Fig.~[XX] in supplementary shows that this multi-layer UVGS mapping can be used to even map real-world complex scenes to structured UV maps and can be reconstructed back with high-fidelity.}

\vspace{-0.1cm}
\noindent\textbf{Baselines \& Metrics}: 
% We use the standard metrics to examine the quality of our reconstructions, unconditional and conditional generation experiments. 
To evaluate the quality of reconstructed 3DGS objects from both Super UVGS image and Autoencoder latent space to 3D, we use PSNR and LPIPS~\cite{lpips2018}. 
The aim is to convert the given 3DGS object to UVGS, and then to Super UVGS, and further to Autoencoder's latent space and calculate the metrics from the reconstructions at every step to prove the proposed method doesn't significantly affect the quality of reconstructions, while also providing a structurally meaningful representation that is much compact and easier to use with existing image based models. 
We compare the generational capabilities of our method against various conditional and unconditional SOTA 3D object generation method including the ones using multiview rendering for optimization DiffTF~\cite{difftf2023}, Get3D~\cite{get3d2022}, methods trying to give structural representation to Gaussians, GaussianCube~\cite{gaussiancube2024}, and general purpose SOTA large 3D content generation models like 
 DreamGaussian~\cite{dreamgaussian2023}, LGM~\cite{lgm2025}, and EG3D~\cite{eg3d2022}. 
We also compare the quality of our generation results using FID and KID.

\begin{figure*}[t]
\centering
\includegraphics[width=0.95\linewidth]{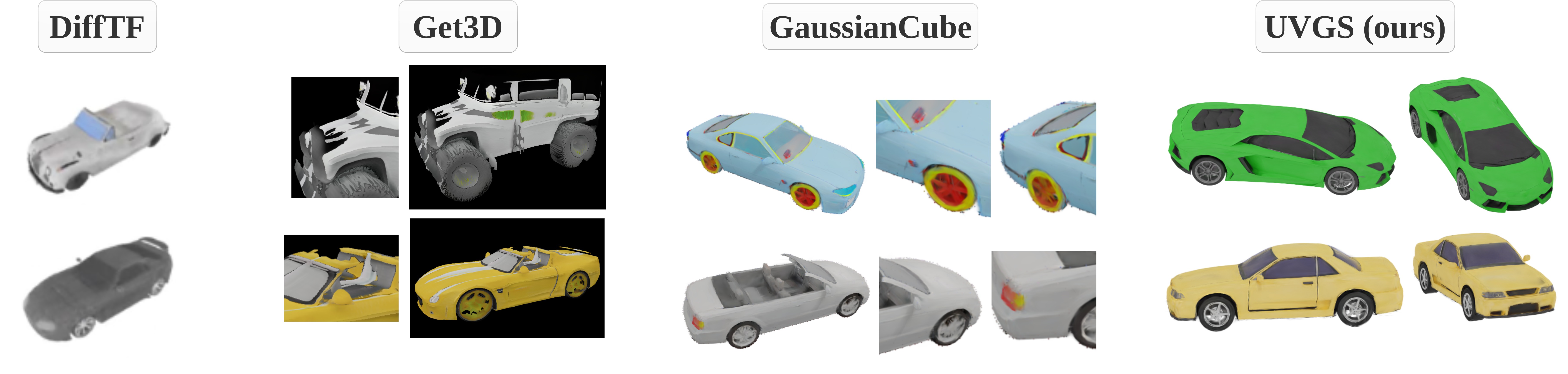} 
\vspace{-0.4cm}
\caption{
Comparison of unconditional 3D asset generation on the cars category with SOTA methods. Figure shows that DiffTF~\cite{difftf2023} produces low-quality, low-resolution cars lacking detail. While Get3D~\cite{get3d2022} achieve higher resolution, it suffers from 3D inconsistency, numerous artifacts, and lacks 3D detail. Similar issues are found in GaussianCube~\cite{gaussiancube2024} along with symmetric inconsistency in the results. In contrast, our method generates high-quality, high-resolution objects that are 3D consistent with sharp and well-defined edges.
}
\vspace{-0.4cm}
\label{fig:compare_cars}
\end{figure*}

% \vspace{0.2cm}

\noindent\textbf{Mapping Network Training Details}
We train the forward and inverse mapping networks to project the obtained UVGS maps $U \in \mathbb{R}^{M\times N \times14}$ to Super UVGS image $S \in \mathbb{R}^{M\times N \times 3}$, and back to the reconstructed UVGS maps $\hat{U} \in \mathbb{R}^{M\times N \times14}$. We provide an in-depth discussion of all implementation details in the supplementary material. 
% \ar{Moved here from Method section: 
% We used a pretrained image VAE [cite] $\Phi_{ae}$ for some epochs to let the mapping networks learn the VAE compatible Super UVGS images (needs to be rewritten in better words). 

\begin{table}[t]
\centering
\setlength{\tabcolsep}{0.75mm}
\renewcommand{\arraystretch}{1.2}
\caption{PSNR and LPIPS comparison for various reconstruction methods using UVGS and Super UVGS representations on Objaverse Cars and Full datasets. 
AE, VAE, VQVAE are pretrained image based models. $K$ is the number of UVGS layers used. 
We also report the compression \% (CP) compared to the fitted 3DGS.}
\resizebox{0.8\columnwidth}{!}{
\begin{tabular}{l|c|c|c}
  \toprule
 \textbf{Method} & \textbf{PSNR($C / F$)} & \textbf{LPIPS($C / F$)} & \textbf{CP($\%$)} \\
 \midrule
 3DGS &  $34.6 ~/~ 34.2$  & $0.02 ~/~ 0.02 $ & $0$ \\
 UVGS (@K=1) & $31.3 ~/~ 31.1$ &$0.06 ~/~ 0.06 $ & $53.0$ \\
 UVGS (@K=2) & $32.8 ~/~ 31.9$ & $0.04 ~/~ 0.05 $ & $45.6$ \\
 UVGS (@K=4) & $34.2 ~/~ 33.2$ & $0.02 ~/~ 0.03 $ & $33.3$ \\
 Super UVGS (@K=1) & $31.2 ~/~ 31.1 $ & $0.07 ~/~ 0.08 $ & $89.7$ \\
 AE (@K=1) & $30.9 ~/~ 30.8 $ & $0.07 ~/~ 0.09 $ & $99.5$ \\
 VAE (@K=1) & $30.6 ~/~ 30.9 $ & $0.07 ~/~ 0.09 $ & $99.5$ \\
 VQVAE (@K=1) & $30.3 ~/~ 30.1 $ & $0.08 ~/~ 0.10 $ & $99.7$ \\
\bottomrule 
\end{tabular}}
\label{table:psnr}
\vspace{-0.5cm}
\end{table}

\subsection{UVGS AutoEncoder and 3DGS Compression}\label{exp:uvgs_ae}
The obtained Super UVGS image is a structurally meaningful representation that can have various applications in the generation and reconstruction of new 3D assets as it contains features that can be learned by the existing image based models.
Through our experiments, we show that a 3-channel Super UVGS image can be directly reconstructed using a pretrained image based Autoencoders or VAEs without any fine-tuning. 
We tested on three different models including image AE, KL-VAE~\cite{KLvae2013}, VQVAE~\cite{vqvae2017} and each performed quite well without any significant quality loss. The reconstruction PSNR and LPIPS values are presented in Table~\ref{table:psnr}.
This means we can now leverage the powerful compression capabilities of image based Autoencoders to compress the storage requirements of 3DGS by more than $99\%$. We have shows the storage comparison results in Table~\ref{table:psnr}.
It is interesting to note that the Super UVGS representation itself can be used to compress the memory requirement for storing 3DGS object by up to $89.7\%$.

% \dilin{we probably don't have enough time, but i guess reviewers would ask: 1/ what if you further finetune the AE/VAE; 2/ what if you train with UVGS from scratch (without even compressing on the channel dimension first}
% \ar{I'm training a VAE from scratch. Hope we get results in the next two days. Otherwise supplementarty or rebuttal.}

% 3DGS Compression

\begin{figure}[t]
\centering
\includegraphics[width=\columnwidth]{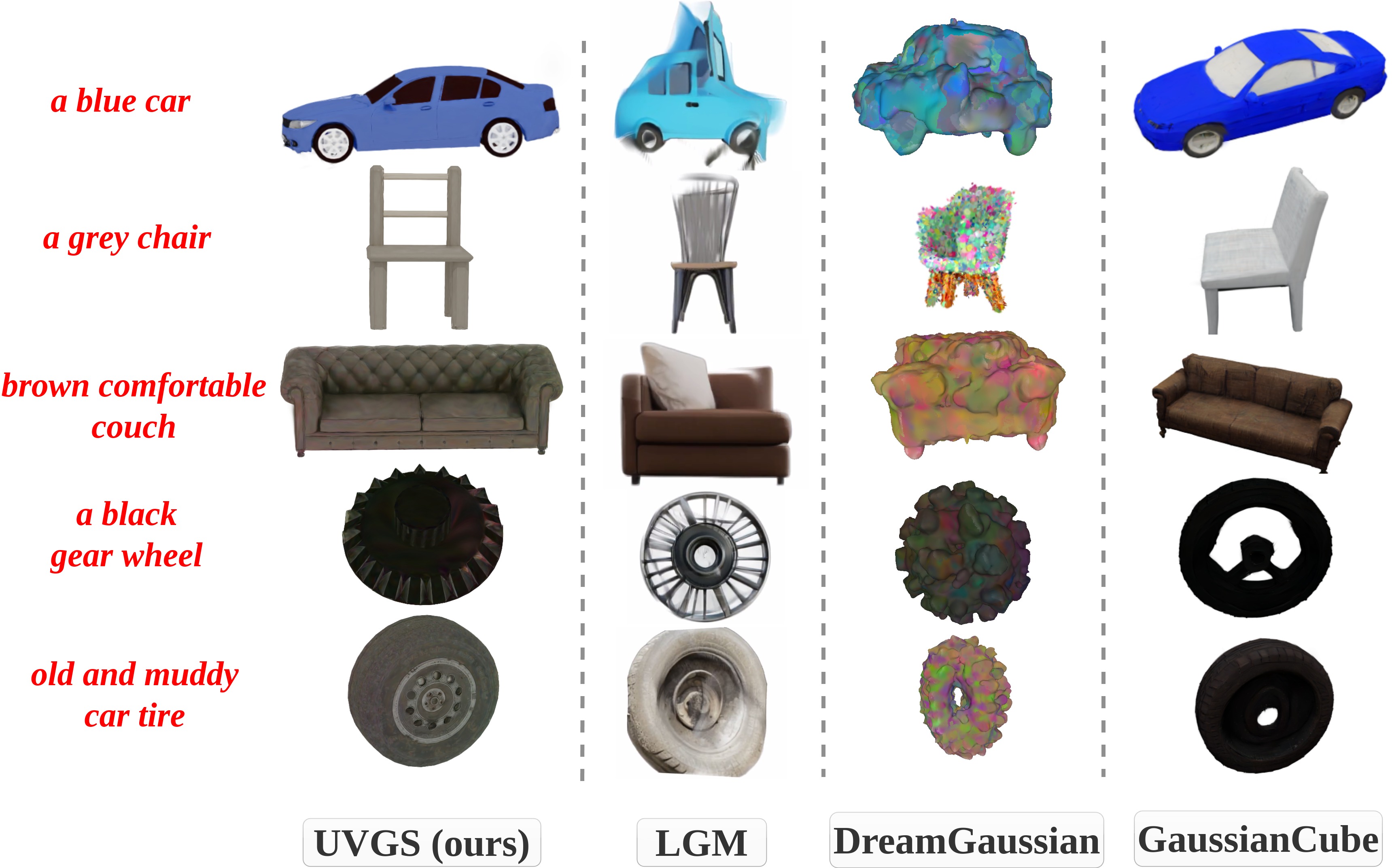} 
\vspace{-0.4cm}
\caption{
We compare the performance of our model against various SOTA methods for text-conditional object synthesis. Our method not only generates high-quality assets for simpler objects, but also for complicated objects with intricate geometry. 
% \srinath{add circle around part of objects that people should focus on.}
}
\label{fig:compare_text}
\vspace{-0.6cm}
\end{figure}

\vspace{-0.1cm}
\subsection{Unconditional \& Conditional Generation}
We aim to show the effectiveness of Super UVGS representation for directly generating 3DGS objects from a learned latent space. 
We consider Super UVGS images as a compact and structured proxy for representing 3DGS objects as it maintains the 3D object information while also providing learnable features. 
The existing methods fail to directly generate a large number of Gaussians (\eg 100K+) to represent objects with sufficient quality either due to the lack of 3D generative model architectures that support such large number of unstructured points, or due to the lack of a large 3DGS dataset \cite{splatterimage2024, lgm2025, triplanemeetsgs2024, gaussiancube2024, gsd2024}. 
We leverage the Super UVGS representation and use the existing 2D image generative models like Diffusion Models~\cite{ddim} for this task. 
Specifically, to train a generative model capable of randomly sampling new high-quality 3DGS assets for various downstream tasks, we use an unconditional Latent Diffusion Model (LDM)~\cite{stable_diffusion} on the obtained Super UVGS images. 
As illustrated in Section~\ref{exp:uvgs_ae}, we can use a pretrained image VAE to map the Super UVGS image to a latent space and reconstruct back. Hence, we only train a LDM on the latent space. More implementation details are provided in the supplementary.

% \vspace{0.2cm}

\noindent\textbf{Unconditional LDM:}
To design a generative model capable of randomly sampling new high-quality 3DGS assets for various downstream tasks, we train an unconditional LDM~\cite{stable_diffusion} on the learned Super UVGS images. 
% We can consider Super UVGS images as a compact and structured proxy for representing 3DGS objects as it maintains the 3D information of the object, while also providing local and global features for learning. 
% As illustrated in the previous section, we are able to use pretrained image VAE for mapping the Super UVGS image to a latent space and back to the reconstructed Super UVGS. To this end, we only need to efficiently train a LDM on the latent space. 
Following~\cite{stable_diffusion, controlnet, egosonics}, we use DDIM~\cite{ddim} for faster and consistent sampling with up to 1000 time steps used in the forward diffusion process, and 20 during denoising. 
% Training was done using AdamW optimizer with a learning rate of $1e-4$ for 50 epochs on $8 \times A100 ~(80GB)$ GPUs.
Once trained, the model is used to randomly sample Super UVGS images, resulting in high quality 3DGS assets through inverse mapping. Results are presented in Fig~\ref{fig:unconditional}. We demonstrate that our method inherently learns to generate multiview consistent images due to the powerful Super UVGS representation unlike most prior works using rendering-based losses. 
% Baseline comparison for unconditional generation is presented in Table~\ref{table:unconditional}.

% \vspace{0.2cm}
% Conditional LDM
\noindent\textbf{Conditional LDM}:
Similar to unconditional generation, we also trained a text-conditioned LDM following the SD's~\cite{stable_diffusion, controlnet, egosonics} pipeline and using the predicted text for our dataset.
The trained model can be used to generate high-quality text-conditioned 3DGS assets that are multiview consistent. The results are demonstrated in Fig~\ref{fig:compare_cars}. 
% Iplementation details are provided in the supplementary.

% Closing Statement
The above experiments proves the effectiveness of our proposed Super UVGS representation in 3D object synthesis using widely available 2D image models. It also highlights that this compact 3-channel Super UVGS representation stores not just the spatial correspondence among different pixels, but also the rich 3D information of the objects. This way, we can easily convert a 3D asset generation problem into a 2D image generation problem without the use of any complex 3D architecture to handle large amount of unstructured and permutation invariant 3DGS primitives, and neither relying upon computationally expensive multiview rendering or SDS loss. Baseline comparison for unconditional and conditional generation is presented in Table~\ref{table:unconditional}.

\subsection{3DGS Inpainting}
\begin{figure}[t]
\centering
\includegraphics[width=0.95\columnwidth]{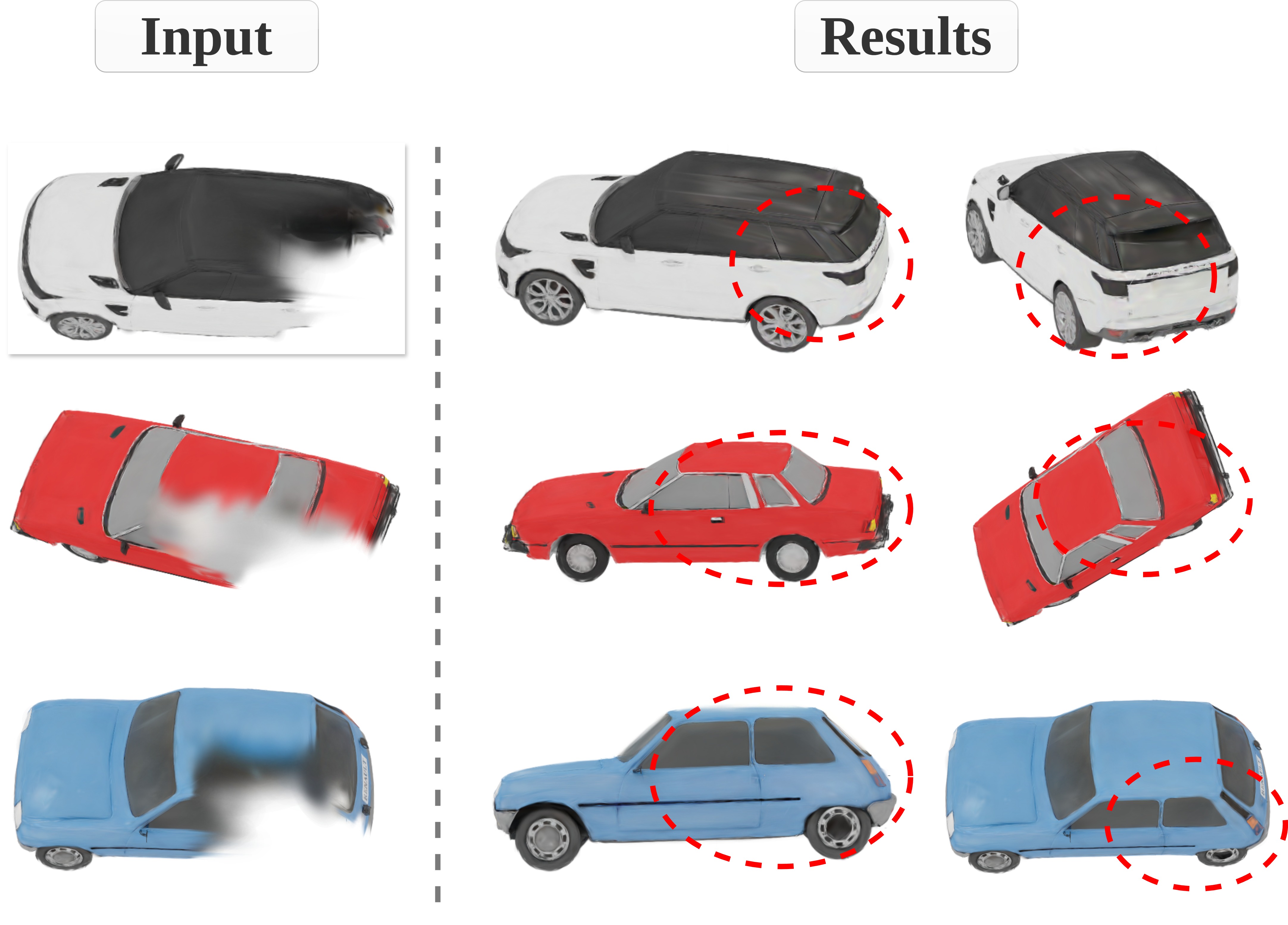} 
\vspace{-0.5cm}
\caption{\textbf{3DGS Inpainting}: We present one of the first inpainting results on 3DGS directly leveraging the Super UVGS images and the denoising capabilities of diffusion models.
% \srinath{add circle around part of objects that people should focus on.}
}
\label{fig:inpaint}
\vspace{-0.4cm}
\end{figure}

Leveraging the powerful Super UVGS representation, we present in Fig.~\ref{fig:inpaint} one of the first experiments on inpainting 3DGS directly without using any multiview rendering or distilling information from diffusion models. 
We try to recover the missing Gaussians by leveraging the denoising capabilities of LDM and trying to predict the missing corresponding parts of the Super UVGS image. We believe, this can have potential applications in sparse view reconstruction.
More details are given in the supplementary.

\subsection{Ablation Studies \& Discussion}

We conduct exhaustive ablation studies to justify some of our framework's design choices including the effect of branching in mapping networks, the use of single layer UVGS maps, and the resolution of UVGS maps. We performed our experiments on our custom Objaverse~\cite{objaverse2023} 3DGS dataset and evaluate the performance of our model in terms of PSNR, SSIM, and LPIPS. The results are presented in Table~\ref{table:ablation}. From the table, it can be seen that by using four layers of UVGS maps (K=4), we can almost match the reconstruction quality of fitted 3DGS results, while we realized that simply with a single layer UVGS, we are able to maintain the overall geometry and appearance of the object for our dataset with a PSNR of more than 30. 
We also compared the reconstruction performance of our method with and without using branching in the mapping network. It can be clearly seen that using branching network significantly increases the reconstruction quality from Super UVGS space. The main reasoning behind this is the specialization of attributes that the branching provides to individually process each attribute first.

\vspace{0.2cm}

\noindent\textbf{Limitations \& Future Work}: 
% Although, our method is able to synthesize and reconstruction a large number of diverse 3D assets using UV mapping, there are some limitations to it. 
While single layer UVGS images can recover the geometry of the object, they sometimes suffer in terms of appearance and the generated objects might look washed out. 
We believe this can be solved by using a multi-layer UVGS maps. 
Similarly, the single layer UVGS map is limited to representing simpler everyday objects, and may not be sufficient to represent highly-detailed and complex objects or scenes. 
In the future, we want to extend this framework to learn features for real-world scenes and complex objects like a human head with multi-layer UV mapping. 
We also want to make this representation more efficient by better utilizing the empty pixels of UVGS maps and Super UVGS images while also maintaining the underlying features and 3D information.
\begin{table}[t]
\centering
\setlength{\tabcolsep}{0.75mm}
\renewcommand{\arraystretch}{1.2}
\vspace{-0.2cm}
\caption{We compare the FID and KID of unconditional generation using the current SOTA methods on 20K randomly generated samples from each method and ours. We also compare our method against SOTA text-conditioned generation frameworks on CLIP Score for 10K generated objects from each method. 
}
\vspace{-0.2cm}
\resizebox{0.85\columnwidth}{!}{
\begin{tabular}{l|cc|||l |c }
 \toprule
 \multicolumn{3}{c}{\textbf{Unconditional Generation}} &  \multicolumn{2}{c}{\textbf{Text-Conditioned Generation}}\\
 \cmidrule(l){1-3} \cmidrule(l){4-5} 
 \textbf{Method} & \textbf{FID} $\downarrow$ & \textbf{KID} $\downarrow$ & \textbf{Method} & \textbf{CLIP Score} $\uparrow$ \\
 \midrule
 Get3D~\cite{get3d2022} &  $53.17$  & $4.19$  & DreamGaussian~\cite{dreamgaussian2023} &  $28.51$  \\
 DiffTF~\cite{difftf2023} & $84.57$  &$8.73$ & Shap. E~\cite{shapeditor2024} & $30.53$  \\
 EG3D~\cite{eg3d2022} & $74.51$ &  $6.62$  &  LGM~\cite{lgm2025} & $30.74$ \\
 GaussianCube & $34.67$ & $3.72$  & GaussianCube~\cite{gaussiancube2024} & $30.34$ \\
\textbf{ UVGS (Ours)} & $\textbf{26.20}$ & $\textbf{3.24}$ &  \textbf{UVGS (Ours)} & $\textbf{32.62}$ \\
 \bottomrule
\end{tabular}}
\label{table:unconditional} 
\vspace{-0.2cm}
\end{table}

\begin{table}[t]
\centering
\setlength{\tabcolsep}{0.75mm}
\renewcommand{\arraystretch}{1.2}
\caption{We present quantitative ablation study for number of UVGS layers (K), UVGS map resolution, and the effect of branching in mapping network on the Objaverse 3DGS dataset.}
\vspace{-0.2cm}
\resizebox{0.85\columnwidth}{!}{
\begin{tabular}{l|cc||l|cc}
 \toprule
 \textbf{Method} & \textbf{PSNR} & \textbf{LPIPS} & \textbf{UVGS Size} & \textbf{PSNR} & \textbf{LPIPS} \\
 \midrule
  UVGS $@K=1$ & $31.1$  & $0.06$ & $512\times512~(@K=1)$ &  $31.1$  & $0.08$ \\
 UVGS $@K=2$ & $31.9$ & $0.05$ & $256\times256~(@K=1)$ &  $28.2$  & $0.23$ \\
 UVGS $@K=4$ & $33.2$ & $0.03$ & Without Branching &  $27.8$  & $0.31$ \\
 \bottomrule
\end{tabular}}
\label{table:ablation} 
\vspace{-0.4cm}
\end{table}

% \begin{table}[t]
% \centering
% \caption{Ablation Study}
% \scalebox{0.8}{\begin{tabular}{c|c|c}
%  \textbf{Method} & \textbf{PSNR} & \textbf{LPIPS} \\
%  \hline
%  \hline
%  UVGS $@K=1$ & $31.1$  & $0.06$ \\
%  UVGS $@K=2$ & $31.9$ & $0.05$ \\
%  UVGS $@K=4$ & $33.2$ & $0.03$ \\
%  \hline

%  UVGS Size &  $-$  & $-$ \\
%  $512\times512~(@K=1)$ &  $31.1$  & $0.08$ \\
%  $256\times256~(@K=1)$ &  $28.2$  & $0.23$ \\
%  Without Branching &  $25.4$  & $-$ \\

% \label{table:ablation} 
% \end{tabular}}
% \end{table}
\vspace{-0.2cm}
\section{Conclusion}
\vspace{-0.2cm}
We introduced a novel method to solve the underlying issues with 3D Gaussian Splatting (3DGS) that prevent the direct integration of them with the large number of existing image foundational models.
% like Autoencoder and Generative models. 
We proposed UVGS - a structured representation for 3DGS obtained by spherical mapping of 3DGS primitives to UV maps. We further squeezed the multi-attribute UVGS maps to a 3-channel unified and structured Super UVGS image, which not only maintains the 3D structural information of the object, but also provides a compact feature space for 3DGS attributes. The obtained Super UVGS images are directly integrated with the existing image foundational models for 3DGS compression and unconditional and conditional generation using diffusion models. Leveraging these Super UVGS images, we showed one of the first inpainting experiments on 3DGS.

\vspace{-0.4cm}
\paragraph{ACKNOWLEDGEMENTS}
 A part of this work was supported by NSF CAREER grant 2143576 and ONR DURIP grant N00014-23-1-2804.

\clearpage
% \newpage
{
    \small
    \bibliographystyle{ieeenat_fullname}
    \bibliography{main}
}

% \clearpage
% \setcounter{page}{1}
\maketitlesupplementary
\setcounter{section}{0}
\setcounter{figure}{0}
\noindent Our supplementary material contains a wide range of information that cover implementation details for our networks and training procedures, as well as a large variety of qualitative results. 

\noindent\textbf{Supplementary Video}: We refer the interested reader to the supplementary video where we provide an overview of how our proposed approach works as well as a plethora of qualitative results across different tasks. 

\section{Spherical Mapping}

\noindent\textbf{Spherical Mapping}: 
Spherical mapping~\cite{sphericalmapping2006} is a fundamental technique in computer graphics that is used to project 3D meshes onto a 2D map generally for texture mapping, where a 2D image is wrapped around a 3D object, such as a cylinder or a sphere. 
However, cylindrical mapping fails to capture the top and bottom parts of the object in the same UV map, and can introduce distortions for objects that extend far in the Z-direction. Hence we opted for spherical mapping the process of which involves converting 3D Cartesian coordinates ($x,y,z$) into spherical coordinates ($\rho, \theta, \phi$) and then mapping these onto a 2D plane. Algorithm~[1] explains spherical unwrapping in detail for a single layer(K=1). The same process can be repeated for multiple layers, by keeping a track of opacity values.

\paragraph{Thresholding Opacity} 3DGS use multiple points with varying opacity values to represent an object from any specific viewpoint. 
However, it is oftentimes noticed that many of these points have very low opacity values and do not contribute to the object's overall representation or appearance.
We filter these points using a threshold opacity value with no impact on the object's overall geometry and representation to reduce the number of tractable primitives.

\begin{algorithm}
\label{algo:spherical}
\caption{Spherical Unwrapping for UVGS map (K=1).}
\begin{algorithmic}[1]
\Require $3DGS \in \mathbb{R}^{n \times 14}$,~ $(M,N) \in \mathbb{Z},~ K=1$
\Ensure $position(\sigma), color(c), scale(s) \in \mathbb{R}^{n \times 3}$
\Ensure $rotation(r) \in \mathbb{R}^{n \times 4},~ opacity(o) \in \mathbb{R}^{n \times 1}$
\State Extract $xyz(\sigma),~opac(o)$ from $3DGS$
% \State $x \gets z1$, $y \gets y1$, $z \gets x1$
\State $ \text{Spherical radius, }r \gets \sqrt{x^2 + y^2 + z^2}$
\State $ \text{Azimuthal Angle, } \theta \gets tan^{-1}(y, x)$
\State $ \text{Polar Angle, } \phi \gets cos^{-1}(z,r)$
\State $(\theta,~ \phi) \gets (\text{deg}(\theta) + 180,~ \text{deg}(\phi))$
% \State $\phi \gets \text{degrees}(\phi)$
\State $\theta_{UV} \gets \text{round}((\theta / 360) \times M)$
\State $\phi_{UV} \gets \text{round}((\phi / 180) \times N)$
\State Initialize $UV_{map} \gets \text{zeros}(M, N, 14)$
\State Initialize $UV_{opac} \gets \text{zeros}(height, width)$
\ForAll{$(t, P, xyz, o)$ in $(\theta_{UV}, \phi_{UV}, 3DGS, opac)$}
    \If{$0 \leq P < height$ and $0 \leq t < width$}
        \If{$UV_{map}[P, t] \text{ is } 0$}
            \State $UV_{map}[P, t] \gets 3DGS[ind]$
            \State $UV_{opac}[P, t] \gets o$
        \Else
            \If{$o > UV_{opac}[P, t]$}
                \State $UV_{map}[P, t] \gets 3DGS[ind]$
            \EndIf
        \EndIf
    \EndIf
\EndFor \\
\Return $UV_{map}$
\end{algorithmic}
\end{algorithm}

\begin{figure*}[t]
\centering
\includegraphics[width=6.4in]{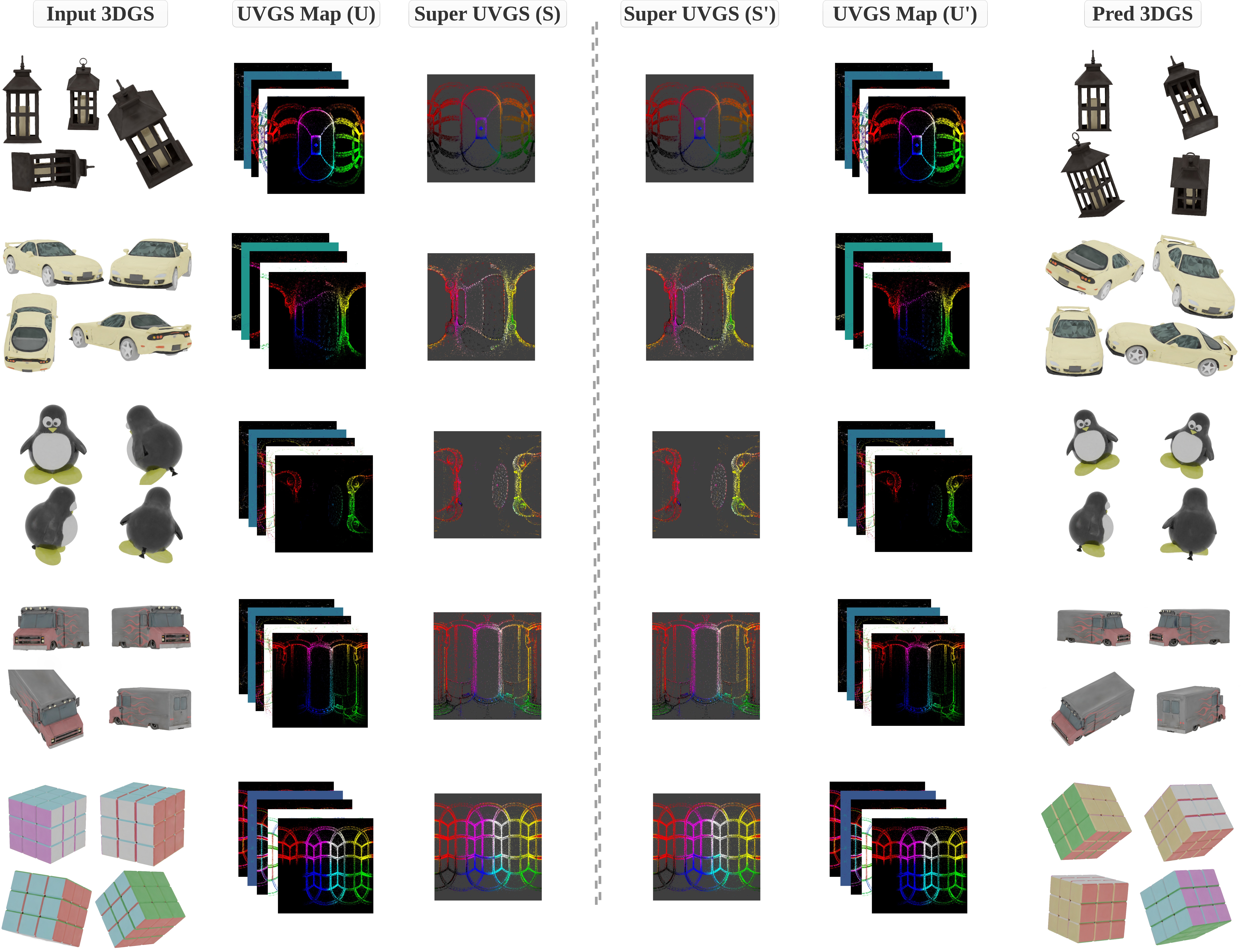} 
\caption{
In this figure, we show the qualitative results of reconstructing 3DGS object using pretrained Image Autoencoder (A) via Super UVGS. We obtain UVGS maps (U) through spherical projection of 3DGS objects, followed by using forward mapping network to get Super UVGS (S). A pretrained AE is used to reconstruct Super UVGS (S'), which can be converted to UVGS maps (U') through inverse mapping network. At last, through inverse spherical mapping, we can get predicted 3DGS object which has the same appearance and geometry as the input object with minimal loss.
}
\label{fig:supp_recons}
\end{figure*}

\begin{figure*}[t]
\centering
\includegraphics[width=6.8in]{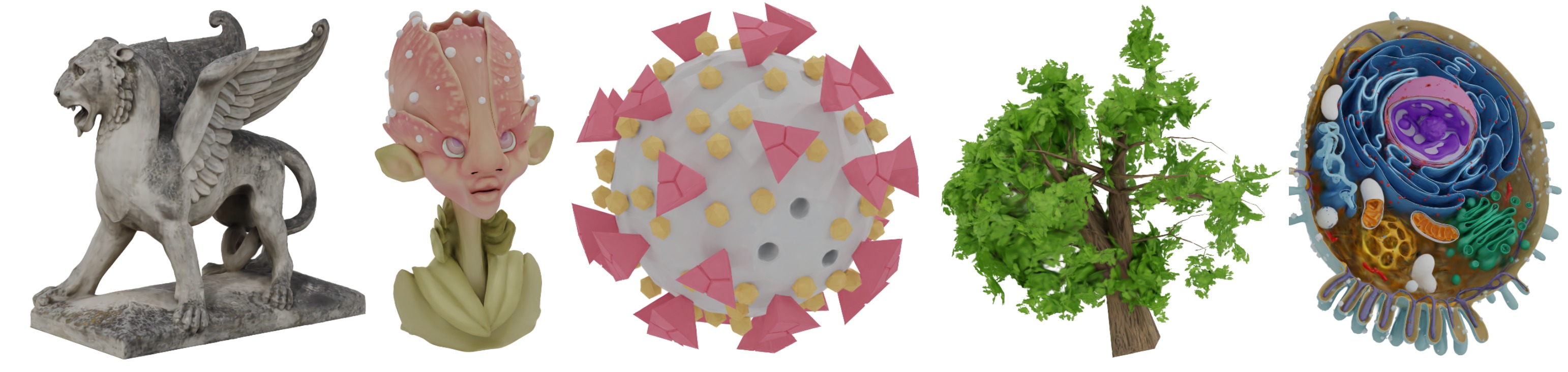} 
\vspace{-0.3cm}
\caption{
Complex object reconstructions (K=4) using pretrained image-based autoencoder.
}
\vspace{-0.6cm}
\label{fig:supp_complex_recons}
\end{figure*}

\paragraph{Dynamic GS Selection and Multiple Layers}
When projecting 3DGS points to UV maps using spherical mapping, multiple points may map to the same pixel in UV space as shown in Fig.~\ref{fig:dynamic_sel}. 
The two 3DGS points $( g_1 )$ and $( g_2 )$ map to the same pixel on UV map $( P_a )$ causing many-to-one mapping. 
However, the UV map can only hold a single 3DGS primitive at any given pixel.
To address this, we propose a Dynamic Selection approach where each UV pixel retains the 3DGS attributes with the highest opacity intersecting the same ray from the centroid to the farthest 3DGS primitive along the ray.
Using the same example in Fig.~\ref{fig:dynamic_sel}, if opacity $o_1$ of Gaussian $g_1$ is less than opacity $o_2$ of $g_2$. then only $g_2$ attributes will be stored in the UV map at pixel $P_a$. 
Through multiple testing, we observed that this method helps retain the overall geometry and appearance of the 3DGS object while resolving many-to-one mapping issues with minimal quality loss.

\begin{figure*}[t]
\centering
\includegraphics[width=6.4in]{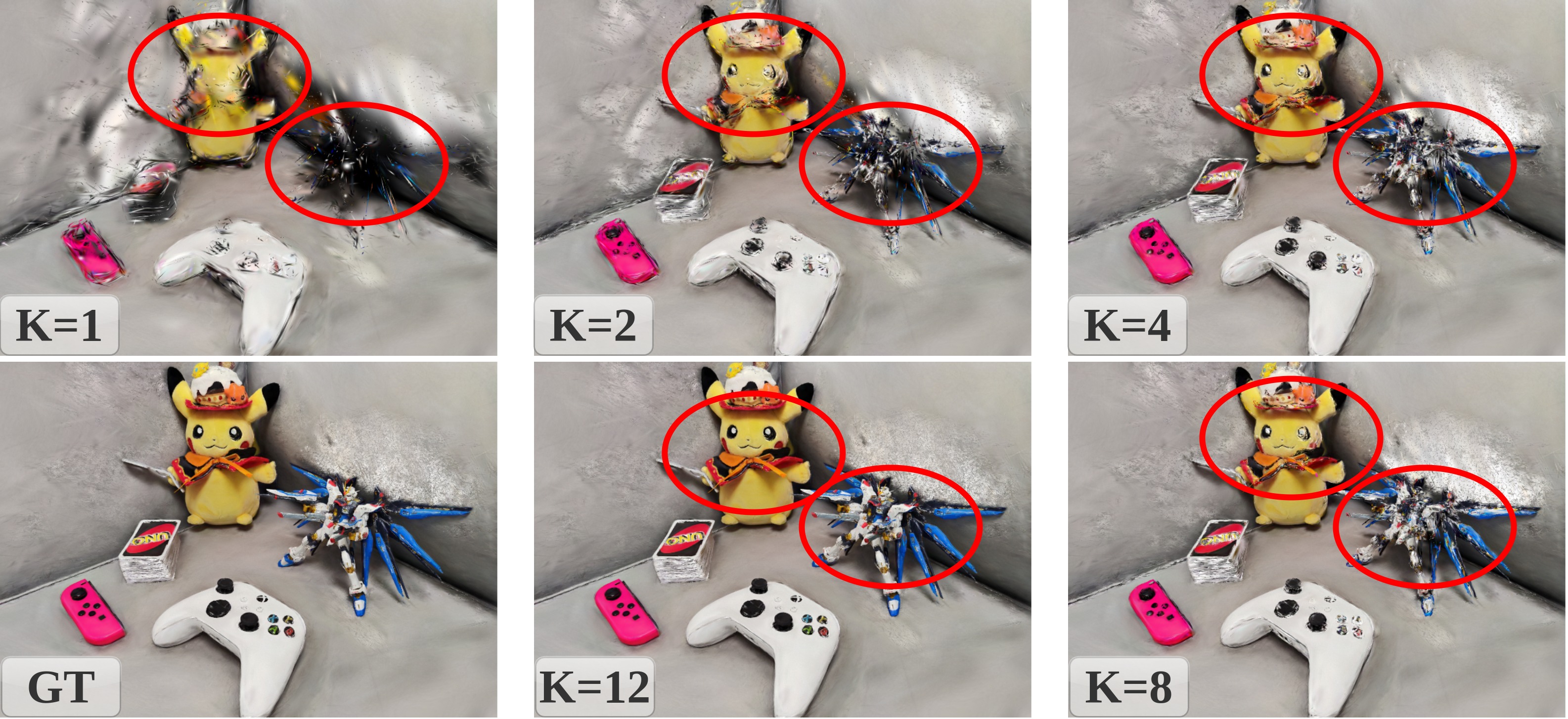} 
\caption{
Reconstruction of a real-world scene for different K values. Smaller K results in many-to-one issue, hence lacking details.
}
\label{fig:supp_recons_K}
\end{figure*}

\begin{figure*}[t]
\centering
\includegraphics[width=5.3in]{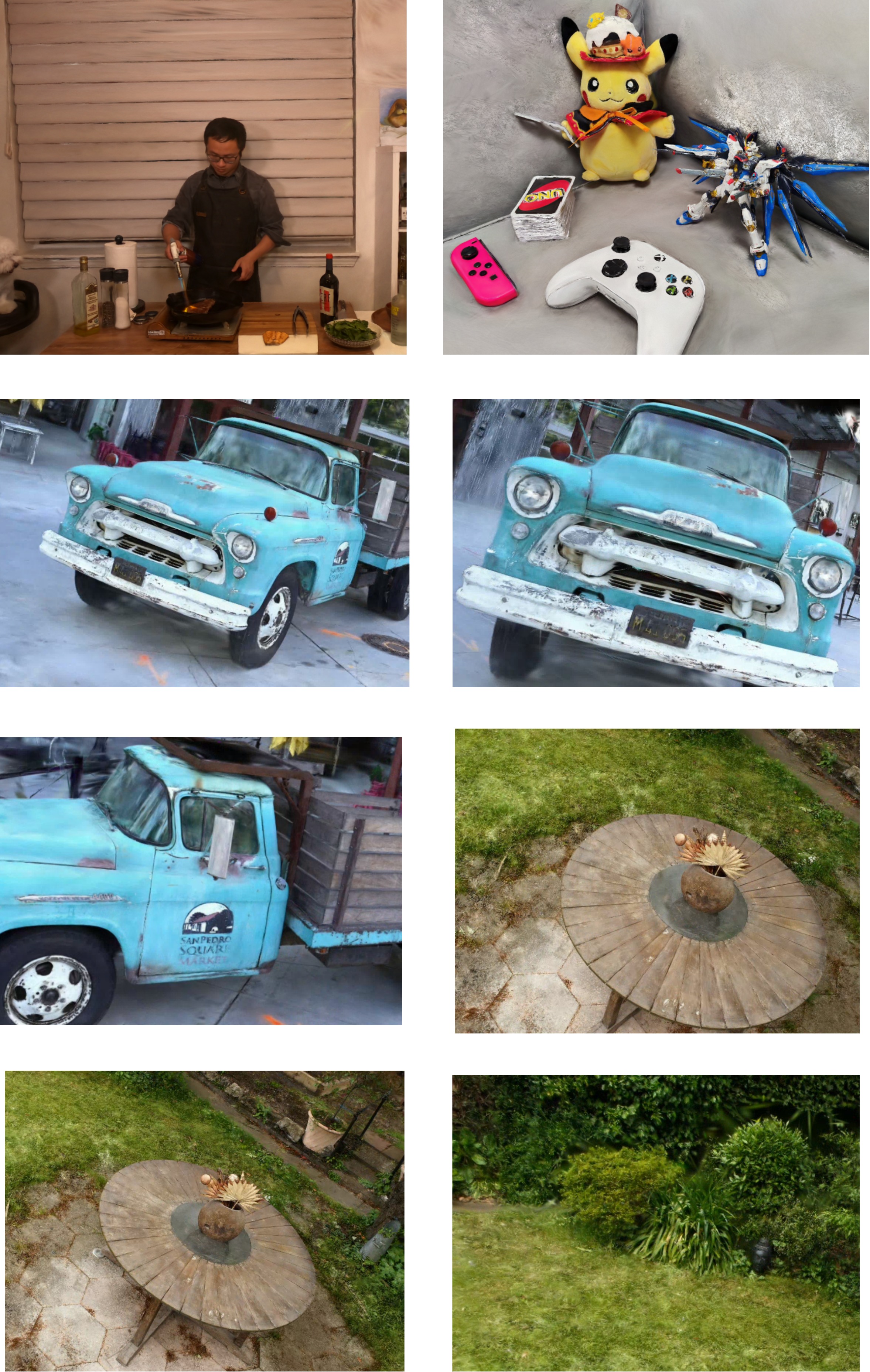} 
\caption{
To show the effectiveness of proposed UVGS maps in capturing the intricacies of a complex real-world scene, we used a 12 layer UV map to reconstruct the 3D scenes.
}
\label{fig:supp_scene}
\end{figure*}

This method with single layer is applicable to most of the objects in our dataset. However, this might fail in the case of more complex objects or real-world scene representation. There could be multiple layers of Gaussians holding higher opacity and contributing to the overall scene's appearance or geometry, and even partial or full occlusions.
To better represent such objects and scenes and to prove the effectiveness of UVGS, we stack multiple layers of UV maps, where each UVGS layer holds the 3DGS primitives of the top-$K^{th}$ opacity value intersecting the same ray. 
This can be accomplished by inscribing the 3DGS object inside multiple spheres where each sphere maps the 3DGS attribute corresponding to the top-$K^{th}$ opacity value along the same ray. 
To show the effectiveness of proposed UVGS maps in capturing the intricacies of a complex real-world scene, we use a 12 layer UVGS map to reconstruct the real-world 3D scenes. The results are presented in Fig.~\ref{fig:supp_scene}. We also compare the effect of increasing the number of UVGS layers in representing a real-world 3D scene in Fig.~\ref{fig:supp_recons_K}
% Fig.~\ref{}\ar{FIG. REFERENCE} in shows that this multi-layer UVGS mapping can be used to even map real-world complex scenes to structured UV maps and can be reconstructed back with high-fidelity. 
In future work, we want to extend this ability for potentially many applications in 3D dynamic scene reconstructions using video diffusion models, and the segmentation or tracking of objects in 3DGS scenes as the features in the UVGS maps can be easily processed with the neural networks and tracked over time. 

% \ns{lets not use concerned anywhere in the paper :) you can write general purpose objects, objects in our dataset or something more descriptive}
% \dilin{i feel the following is not necessary..it's very clear before already.} \st{Note that a multi-layer UVGS map should not be confused with a 14-channel UVGS $U \in \mathbb{R}^{M\times N \times14}$ representing the 5 attributes. A 2-layer UVGS map will have \(14 \times 2 = 28\) channels, with each attribute represented by 2 identical attribute specific layers. 
% Similarly, a 4-layer UVGS map will have \(14 \times 4 = 56\) channels.}

\section{Mapping Networks}

\noindent\textbf{Forward Mapping Details}:
This process is defined as:
\begin{equation}
   f^f_{map} = [ ~[\phi^f_{P}(\sigma)] ~[\phi^f_{T}([r,s])] ~[\phi^f_{A}[o,c]] ~] 
\end{equation}
   
The central branch ($\phi^f_{C}$) is composed of $2L$ hidden Convolution layers.
The first $L$ hidden convolution layers increase the feature dimension at each step, while the last $L$ layers does the inverse and squeezes the high-dimensional feature maps to 3 channels to output Super UVGS image $S \in \mathbb{R}^{M\times N \times 3}$. 
% \dilin{no need to highlight all the dimensions and details here? or write it in a tone that suggests one particular implementation, as there are many other design approaches.}
% \ns{Same thing with before. Explain the intuition behind this choices and give me insights to understand why this makes sense.} \ar{Justufucation at the end of this section}
\begin{equation}
    S = tanh(~\phi^f_{C}[f^f_{map}]~) ~\in~\mathbb{R}^{(H,W,3)}
\end{equation}
Each CNN layer is followed by a batch normalization layer and ReLU activation both in multi-branch and central branch modules. 
The last layer of central branch is activated using $tanh$ to ensure the Super UVGS doesn't take any ambiguous value resulting in gradient explosion or undesired artifacts. 
The obtained Super UVGS $S$ representation squeezes all the 3DGS attributes to a 3 dimensional image while also maintaining local and global structural correspondence among them.

\begin{figure*}[t]
\centering
\includegraphics[width=6.4in]{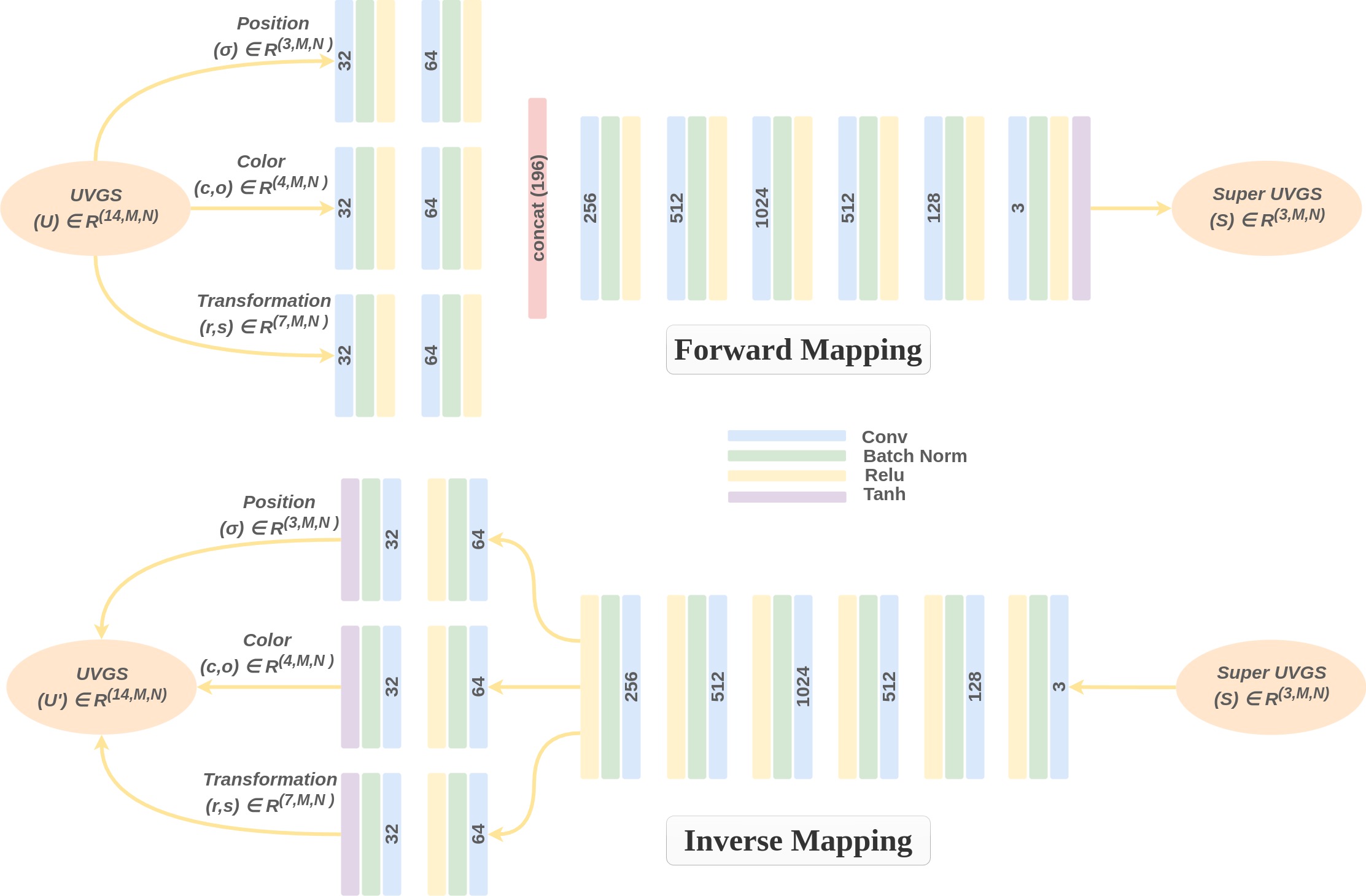} 
\caption{
Forward Mapping Network for UVGS to Super UVGS mapping. The inverse mapping network follows just the inverse of this architecture with each attribute-specific branch now followed by $tanh()$ at the end.
}
\label{fig:mapping_network}
\end{figure*}

\noindent\textbf{Inverse Mapping}:
The first $L$ layers in the Central branch increases the feature dimension and the last $L$ layers reduces them to obtain a combined feature map. 

    \[
   f^i_{map} = \phi^i_{C}(S) 
   \]

The final layer is a set of 3 branches projecting the features to position, translation, and appearance attributes, respectively. 
% \dilin{similar issue, no need to use exact numbers, espeically in equations} 
% \ar{ToDo: Remove numbers from all equations.}

    \[
   f^i_{\sigma} = [\phi^i_{P}(f^i_{map})] 
   \]
    \[
   f^i_{r,s} = [\phi^i_{T}(f^i_{map})] 
   \]
    \[
   f^i_{o,c} = [\phi^i_{A}(f^i_{map})] 
   \]

Similar to the forward mapping network, each layer in the central branch and attribute specific branches is followed by batch normalization and $Relu(.)$ activation. 
The last set of branch layers are activated using $tanh(.)$ to prevent ambiguous values resulting in gradient explosion or reconstruction artifacts.

   \[
   \hat{U} = tanh(~ [ [f^i_{\sigma}]~[f^i_{r,s}]~[f^i_{o,c}] ] ~)
   \]

\noindent \textbf{Losses Details}: We used MSE to focus on pixel-wise difference during the training. We solely used MSE for a few iterations to make the mapping networks learn the overall structural representation of the UVGS map using:
\begin{equation}
\mathcal{L}_{mse} = \frac{1}{n} \sum_{i=1}^n (U_i - \hat{U}_i)^2.
\end{equation}
After training the model for few iterations using MSE, we introduce the LPIPS loss giving same weight to both MSE and LPIPS over a few iterations. We observed that increasing the weight value of LPIPS over the iterations resulted in better and faster convergence results.
\begin{equation}
    \mathcal{L}_{lpips} = \sum_{l} w_l \left\| \phi_l(x) - \phi_l(y) \right\|^2,
\end{equation}
where $\phi_l(x)$ and $\phi_l(y)$ are feature maps extracted from pretrained layers of AlexNet\cite{lpips2018}.

\noindent \textbf{Mapping Training Details}: 
% We fix the number of hidden layers in Central branch of forward and inverse mapping networks to six. The first three layers increase the feature dimension at each step reaching a maximum of 1024 features. While the last 3 layers does the inverse and squeezes the 1024 dimensional feature maps to 3 channels to output Super UVGS image $S \in \mathbb{R}^{M\times N \times 3}$. We used two layers in each of the attribute specific branch in both forward and reverse mapping networks.
% \ns{We already mentioned all the below stuff above no? I don't think the below text fits well in "experiments" and I'd suggest moving it to Suppl eventually}
% However, it should be noted that each set of attributes in 3DGS representing the mean position, transformation, and color has a completely different distribution of values and poses a significant burden on the CNN if processed together leading to uncanny gradient values and slower convergence. Thus, we propose to use a multi-branch network where the attribute specific branches implicitly learns to process these different attributes focusing on their individual features before passing them to the central branch. The central branch obtains a concatenated stack of processed attributes and exploits the correlation between different attributes by extracting the local feature correspondence between them and maps them to a 3 channel Super UVGS image. 
Before training the models, we normalized the different attributes in UVGS to [$-1,1$] using the same normalization functions as used in 3DGS paper\cite{3dgs2023}. 
The normalized UVGS maps are used to train the multi-branch forward and reverse mapping networks using MSE and LPIPS loss.
We trained the mapping networks on $8 \times A100 ~(80GB)$ GPUs with a Batch Size of 96 for 120 hours using Adam optimizer with a learning rate of $6e-5$ and set $\beta_1 = 0.5$ and $\beta_2 = 0.9$ with weight decay of $0.01$. 
We set the $\lambda$ for LPIPS loss to be $0$ for the first 24 hours of training and gradually increased it from 1 to 10 for the remaining training in a step of 1.

\subsection{Interpolation with UVGS}

\begin{figure}[!h]
\centering
\includegraphics[width=3.2in]{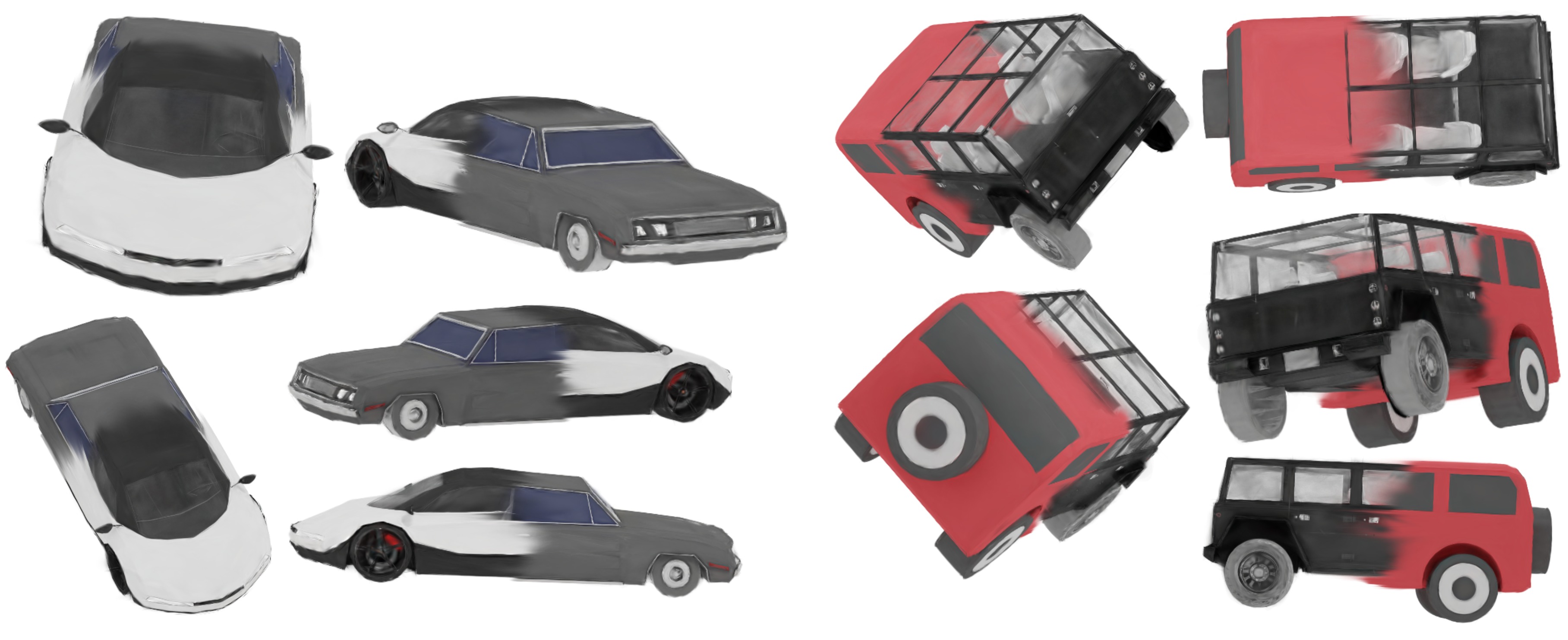} 
\caption{
Linear interpolation between two 3DGS objects using SuperUVGS representation.
}
\label{fig:interpolation}
\end{figure}

We show that the proposed SuperUVGS representation can be used to perform local editing and interpolation directly in the UV domain. We can perform edits like swapping the parts of one object from the other, cropping the 3D object, or merging two objects together simply with the SuperUVGS images without any learning based method. The results are demonstrated in Fig~\ref{fig:interpolation}.

\section{LDM - Unconditional and Conditional Generation}

\noindent\textbf{Caption Generation} To generate the relevant text captions for the objects in our dataset for conditional generation, we leverage CLIP~\cite{clip}, BLIP2~\cite{blip22023}, and GPT4~\cite{gpt42023} very similar to \cite{cap3d2024}. Specifically, we use BLIP2 to generate $N$ different captions for randomly selected 20 views from the 88 rendered views for each object in the dataset. CLIP encoders are used to encode and calculate the cosine similarity between the $N$ generated caption per view and the corresponding 20 views. The caption with max similarity is assigned to that particular view, resulting in 20 different captions for the same object. We now use GPT4 to extract a single caption distilling all the given 20 descriptions. We found that the resulting captions were very appropriate to the input objects, and thus we directly used them for conditional generation.

% \dilin{maybe move to experiments or even appendix?} 
LDMs~\cite{stable_diffusion, ddim} use pretrained VAEs~\cite{transformer_vae} to convert the original image $x \in R^{H\times W\times 3}$ into a compact latent representation $z \in R^{h\times w\times c}$, where the forward and reverse diffusion processes are applied~\cite{stable_diffusion}. 
The VAE decoder then converts the compact latent representation back to pixels. 
The objective function in latent diffusion model can be written as:

\begin{equation}
    \mathbb{L}_{LDM} := \mathbb{E}_{\epsilon(x),\epsilon \sim \mathbb{N}(0,1), t } [ || \epsilon - \epsilon_\theta ( z_t, t ) ||_2^2 ]
\end{equation}

where, $\mathbb{N}(0,1)$ is the Normal distribution, and $t$ is the number of time steps and $z_t$ is the noisy sample after $t$ time steps.

Training was done using AdamW optimizer with a learning rate of $1e-4$ for 75 epochs on $8 \times A100 ~(80GB)$ GPUs. 

Once trained, we can randomly sample new high-quality 3DGS assets from the learned generative model.

% Note that we do not use any conditioning vector for unconditional generation and instead replace the cross-attention layer with the self-attention layer to train the denoising process without any extra conditioning information.
% \ns{For practical reasons ;) I don't want to write Stable Diffusion as a word}

To allow generation of objects from text, we also trained a conditional LDM, where we use Stable Diffusion (SD)~\cite{stable_diffusion} pipeline as it can use text prompt conditioning to guide the image generation through cross-attention. Similar to unconditional LDM, we use pretrained SD's VAE for mapping the Super UVGS image to a latent space and back to the reconstructed Super UVGS.
The text prompts are given to a pretrained CLIP~\cite{clip} text encoder to generate a text embedding $c_t \in \mathbb{R}^{77 \times 768} $, which is then passed to the UNet encoder of SD for cross-attention. We used a set of CLIP encoder and BLIP2~\cite{blip22023}, and GPT4~\cite{gpt42023} to generate captions for our dataset. 
The overall objective function for conditional LDM now becomes:
\begin{equation}
    \mathbb{L}^C_{LDM} := \mathbb{E}_{\epsilon(x),\epsilon \sim \mathbb{N}(0,1), t, c_t } [ || \epsilon - \epsilon_\theta ( z_t, t, c_t ) ||_2^2 ]
\end{equation}

where, $\epsilon_\theta (\cdot, t)$ is a time-conditional U-Net~\cite{unet} model, $\mathbb{N}(0,1)$ is the Normal distribution, $z_t$ is the latent code, and $c_t$ is the text embedding.
Training was done using AdamW optimizer with a learning rate of $1e-4$ for 50 epochs on $8 \times A100 ~(80GB)$ GPUs.
Once trained, this conditional LDM can use used to generate text-conditioned Super UVGS images, which can later be mapped to high-quality 3DGS objects.

% \section{ShapeNet Experiments}

% To show the effectiveness of our method againt varying datasets, we also conducted experiments on ShapeNet~\cite{shapenet2015} dataset. We fine-tuned the trained Mapping Networks and Unconditional and Conditional Diffusion Models on "cars" category of ShapeNet dataset. The results are presented in Fig.~\ref{fig:supp_uncond}. 

\begin{table}[t]
\centering
\setlength{\tabcolsep}{0.75mm}
\renewcommand{\arraystretch}{1.2}
\caption{We compare the FID and KID of unconditional generation using the current SOTA methods on 20K randomly generated samples from each method and ours. We also compare our method against SOTA text-conditioned generation frameworks on CLIP Score for 10K generated objects from each method. 
}
\resizebox{\columnwidth}{!}{
\begin{tabular}{l|cc|||l |c }
 \toprule
 \multicolumn{3}{c}{\textbf{Unconditional Generation}} &  \multicolumn{2}{c}{\textbf{Text-Conditioned Generation}}\\
 \cmidrule(l){1-3} \cmidrule(l){4-5} 
 \textbf{Method} & \textbf{FID} $\downarrow$ & \textbf{KID} $\downarrow$ & \textbf{Method} & \textbf{CLIP Score} $\uparrow$ \\
 \midrule
 Get3D~\cite{get3d2022} &  $53.17$  & $4.19$  & DreamGaussian~\cite{dreamgaussian2023} &  $28.51$  \\
 DiffTF~\cite{difftf2023} & $84.57$  &$8.73$ & Shap. E~\cite{shapeditor2024} & $30.53$  \\
 EG3D~\cite{eg3d2022} & $74.51$ &  $6.62$  &  LGM~\cite{lgm2025} & $30.74$ \\
 GaussianCube & $34.67$ & $3.72$  & GaussianCube~\cite{gaussiancube2024} & $30.34$ \\
\textbf{ UVGS (Ours)} & $\textbf{26.20}$ & $\textbf{3.24}$ &  \textbf{UVGS (Ours)} & $\textbf{32.62}$ \\
 \bottomrule
\end{tabular}}
\label{table:supp_unconditional} 
\vspace{-0.1cm}
\end{table}

\section{Comparison with Baselines}

We compare the generational capabilities of our method against various conditional and unconditional SOTA 3D object generation method on ShapeNet-cars dataset. Specifically, we used the methods using multiview rendering for optimization, like DiffTF~\cite{difftf2023} and Get3D~\cite{get3d2022}. We also compared our approach again the current SOTA methods trying to give structural representation to Gaussians, including GaussianCube~\cite{gaussiancube2024} and TriplaneGaussian~\cite{triplanemeetsgs2024}. We also compared against general purpose SOTA large 3D content generation models like DreamGaussian~\cite{dreamgaussian2023}, LGM~\cite{lgm2025}, and EG3D~\cite{eg3d2022}. 

\begin{figure*}[!ht]
\centering
\includegraphics[width=6.8in]{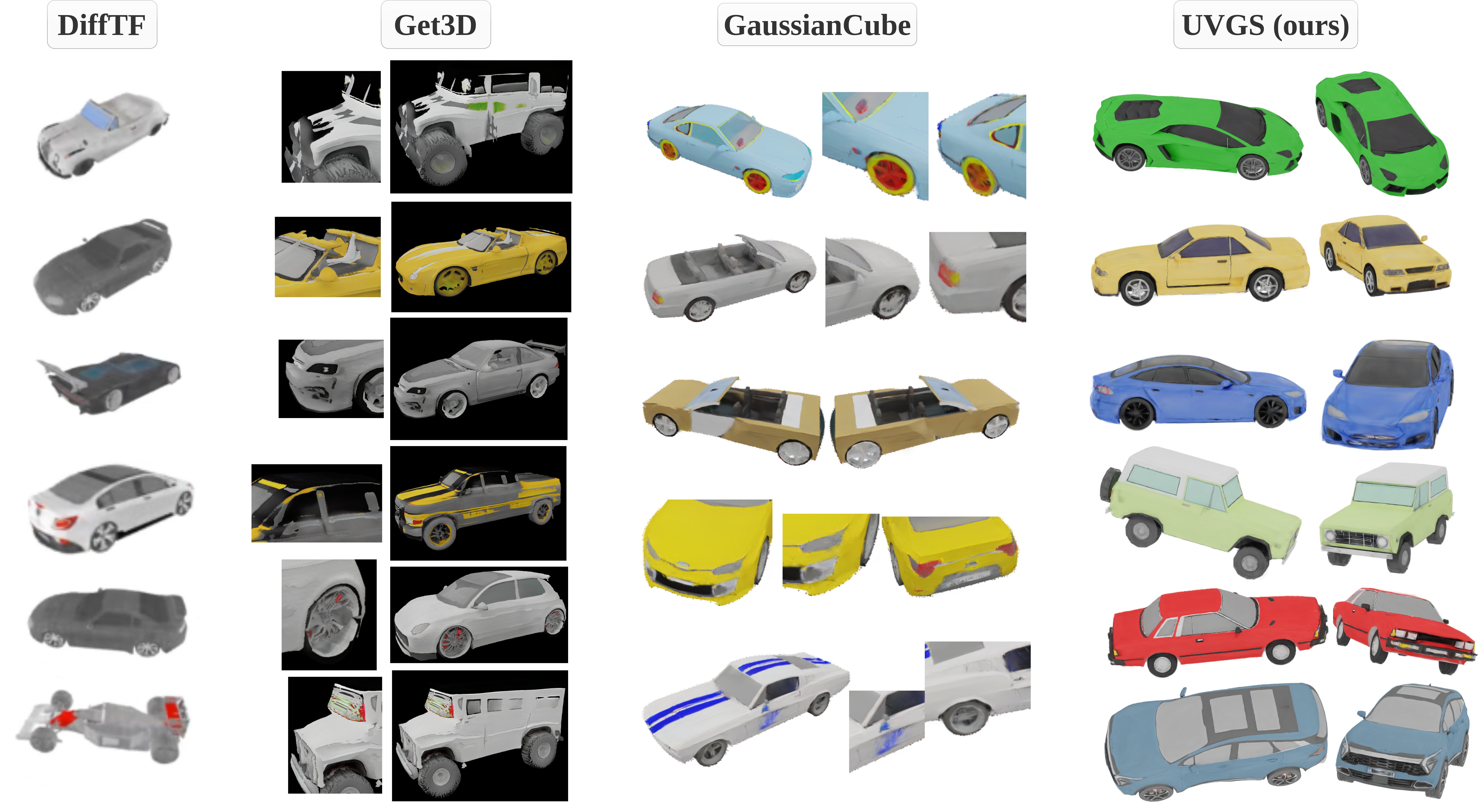} 
\caption{
Here we show more comparison of unconditional 3D asset generation on the cars category with SOTA methods. Figure shows that DiffTF~\cite{difftf2023} produces low-quality, low-resolution cars lacking detail. 
While Get3D~\cite{get3d2022} achieve higher resolution, it suffers from 3D inconsistency, numerous artifacts, and lack richness in 3D detail. Similar issues are found in GaussianCube~\cite{gaussiancube2024} along with symmetric inconsistency in the results. 
In contrast, our method generates high-quality, high-resolution objects that are 3D consistent with sharp, well-defined edges.
The top three rows show the unconditional generation results of our method using ShapeNet dataset, while the bottom 3 show from Objaverse dataset.
}
\label{fig:supp_uncond}
\end{figure*}

\begin{figure*}[!h]
\centering
\includegraphics[width=6.4in]{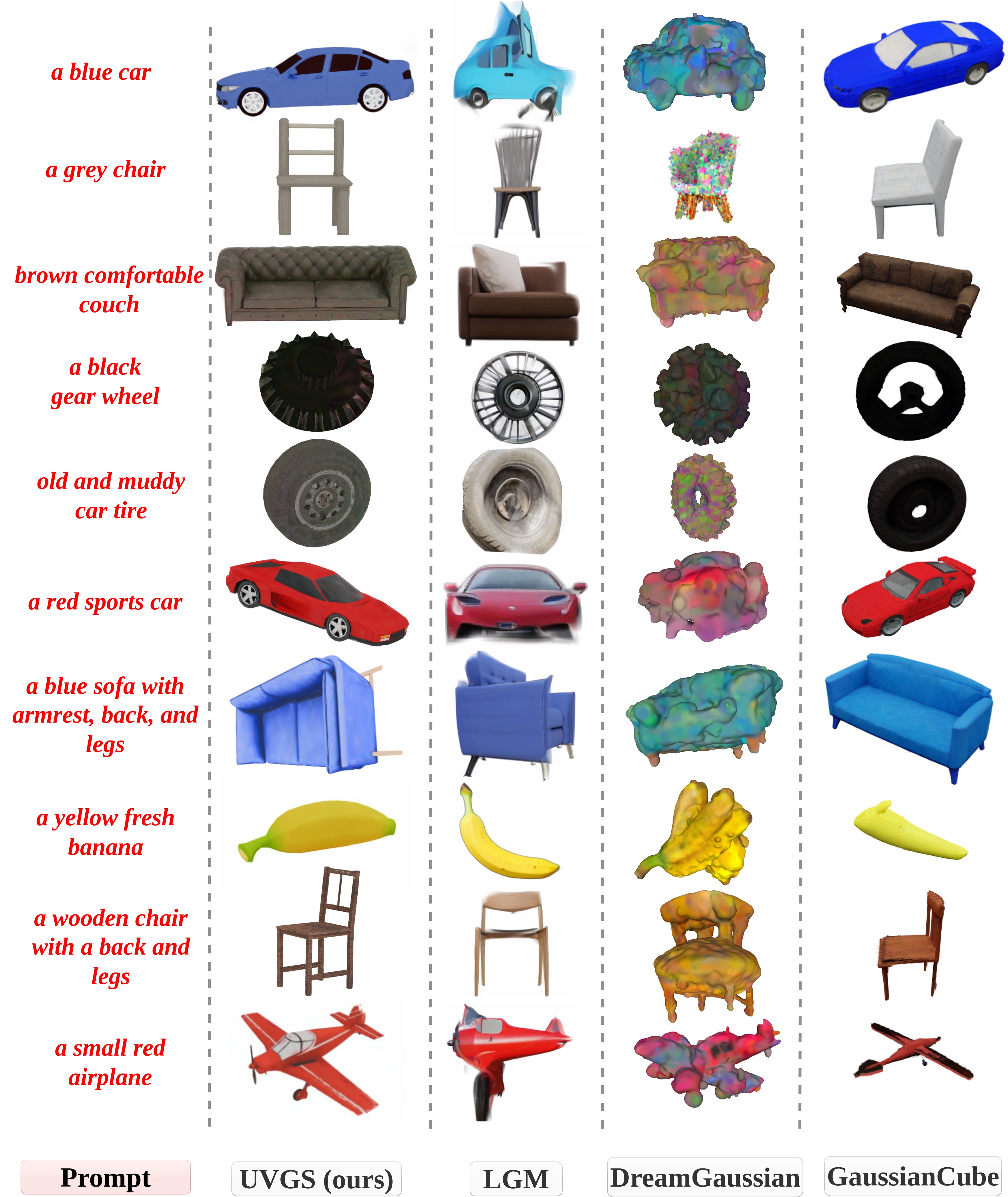} 
\caption{
Text-conditioned generation results on various baselines and the proposed method. Our method not only generates high-quality assets for simpler objects, but also for complicated objects with intricate geometries like \textit{the wheel} or \textit{the airplane}.
}
\label{fig:supp_cond}
\end{figure*}

To compare the quality of our generation results, as a standard practice, we use FID and KID for unconditional generation, and Clip Score for text-conditioned generation. Table~\ref{table:unconditional} quantitatively compares the unconditional and conditional generation results of our method again various SOTA methods. 
% We also compare the results of our fine-tuned ShapeNet cars generation framework and the results are presented in Table~\ref{table:supp_unconditional} as \textbf{UVGS-S}. 
From this table, it can be seen that our method performs a good job in unconditional generation of good quality 3D assets. The main reason behind this is the learned Super UVGS representation which not only maintains the appearance of the 3DGS object, but also serves as a proxy for geometrical shape by encoding all the 3DGS attributes into the same coherent feature space. Table~\ref{table:supp_unconditional} compares the CLIP Score of our text-conditioned generation results and the current SOTA methods. The unconditional and conditional qualitative comparison results are presented in Fig.~\ref{fig:supp_cond} and Fig.~\ref{fig:supp_uncond}, respectively.

\end{document}